\documentclass[journal]{IEEEtran}
\usepackage{amsmath,amsfonts}
\usepackage{algorithmic}
\usepackage{algorithm}
\usepackage{array}
\usepackage[caption=false,font=normalsize,labelfont=sf,textfont=sf]{subfig}
\usepackage{textcomp}
\usepackage{stfloats}
\usepackage{url}
\usepackage{verbatim}
\usepackage{graphicx}
\usepackage{cite}
\usepackage{hyperref}
\usepackage{bbm}

\usepackage{xcolor}
\definecolor{mypink}{RGB}{216, 18, 126}
\definecolor{citeblue}{RGB}{1, 113, 188}

\hypersetup{
    colorlinks=true,
    linkcolor=mypink,
    urlcolor=mypink,
    citecolor=citeblue,
}

\usepackage{booktabs}
\usepackage[dvipsnames,table]{xcolor}  
\usepackage{tikz}            
\usepackage[edges]{forest}   
\definecolor{hidden-draw}{RGB}{0,0,0}
\definecolor{hidden-pink}{rgb}{0.98, 0.94, 0.75}
\definecolor{level0}{rgb}{0.67, 0.88, 0.69}
\definecolor{level1}{rgb}{0.98, 0.92, 0.84}
\definecolor{level2}{rgb}{0.8, 0.8, 1.0}
\definecolor{level3}{rgb}{1.0, 0.71, 0.76}
\definecolor{level4}{rgb}{0.49, 0.99, 0.0}

\usepackage{multirow}
\usepackage{threeparttable}
\usepackage{color}
\usepackage{utfsym}

\definecolor{limegreen}{rgb}{0.2, 0.8, 0.2}

\begin{document}

\title{Mixture-of-Experts for Time Series Analysis: Taxonomy, Progress, and Prospects}

\author{
        Yanru Sun,
        Dilfira Kudrat, 
        Zongxia Xie$^\dag$,
        Yukun Yan,
        Emadeldeen Eldele,
        Qinghua Hu,
        Chee Keong Kwoh,
        Zhenghua Chen,
        Xiaoli Li,~\IEEEmembership{Fellow,~IEEE,}
        and Min Wu$^\dag$

\thanks{Yanru Sun, Dilfira Kudrat, Zongxia Xie, and Qinghua Hu are with Tianjin University. (e-mail: yanrusun@tju.edu.cn, dilfira\_49@tju.edu.cn, caddiexie@hotmail.com, and huqinghua@tju.edu.cn)}
\thanks{Yukun Yan is with Harbin Engineering University. (e-mail: yanyukun@hrbeu.edu.cn)}
\thanks{Emadeldeen Eldele is with the Department of Computer Science, Khalifa University, Abu Dhabi, UAE. (e-mail: emad0002@ntu.edu.sg)}
\thanks{Chee Keong Kwoh is with Nanyang Technological University. (e-mail: ASCKKWOH@ntu.edu.sg)}
\thanks{Zhenghua Chen is with the University of Glasgow. (e-mail: chen0832@e.ntu.edu.sg)}
\thanks{Xiaoli Li is with the Singapore University of Technology and Design. (e-mail: xiaoli\_li@sutd.edu.sg)}
\thanks{Min Wu is with the Institute for Infocomm Research, A*STAR. (e-mail: wumin@a-star.edu.sg)}
\thanks{$^\dag$Corresponding author: Zongxia Xie and Min Wu.}
}

\maketitle

\begin{abstract}
Mixture-of-Experts (MoE) has recently demonstrated remarkable success across diverse applications.
Its ability to decompose data into sub-patterns and assign them to specialized experts not only improves modeling efficiency but also enables computational scalability, making it well-suited for large-scale, heterogeneous, and non-stationary time-series data.
However, unlike in computer vision and natural language processing, where MoE architectures often rely on static or spatial assumptions, time-series data introduces fundamentally different challenges.
It requires routing mechanisms that respond to temporal evolution and experts capable of modeling complex, multi-scale, and cross-variable dependencies.
Without a systematic overview, the field risks repeatedly adopting designs that fail to generalize under dynamic temporal conditions.
To address this gap, we review existing MoE approaches for time-series analysis, propose a unified taxonomy based on \emph{routing strategy} and \emph{expert design}, and analyze their methodological innovations, strengths, and limitations. The taxonomy not only organizes current methods but also exposes recurring design patterns, uncovers underexplored combinations, and clarifies how routing and expert choices jointly shape model behavior in temporal settings.
We then summarize key insights emerging from the literature and outline future directions across three key perspectives: structural design, adaptive intelligence, and trustworthy deployment.
This survey not only organizes recent advances but also provides a clearer foundation for developing efficient, adaptive, and reliable time-series MoE systems.
The project page is available at: \url{https://github.com/syrGitHub/Awesome-MoE4TS}.
\end{abstract}

\begin{IEEEkeywords}
Time series modeling, mixture-of-experts, temporal heterogeneity, adaptive routing, expert specialization.
\end{IEEEkeywords}

\section{Introduction}
\begin{figure}[!t]
\centering
\includegraphics[width=1.0\linewidth]{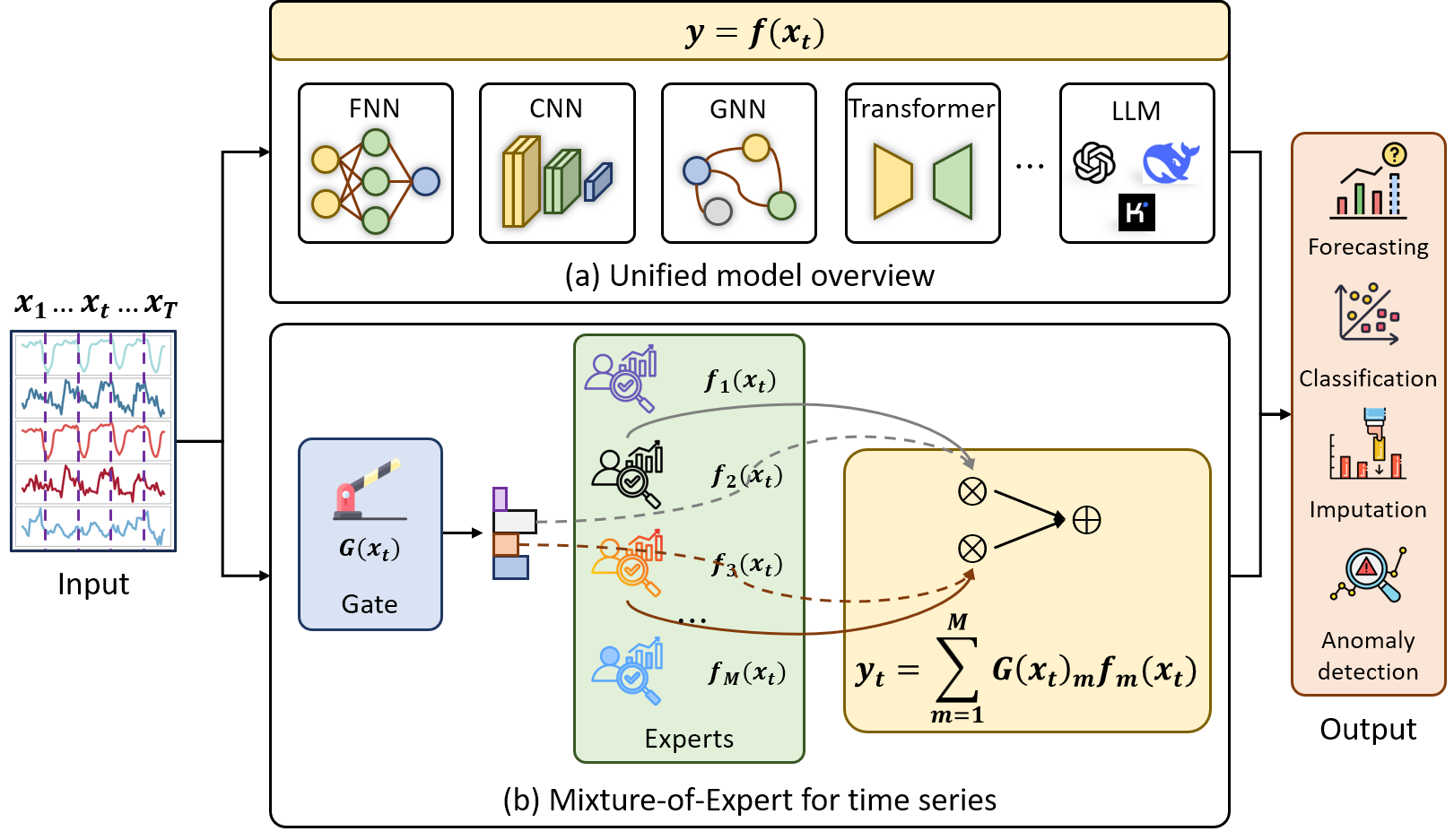}
\caption{Illustration of the difference between a unified model and a mixture-of-experts (MoE) model. The unified model applies a single mapping $y=f(x_t)$ to all inputs, while the MoE model adaptively combines multiple experts through input-dependent gating.}
\label{fig:Unified-MoE}
\end{figure}

\IEEEPARstart{T}{ime} series data are ubiquitous in various real-world scenarios, such as healthcare \cite{eldele2021attention, el2024ecgtransform, jin2025indicators}, transportation \cite{long2024unveiling, kong2024spatio, jin2024spatial}, manufacturing \cite{eldele2025unifault, ragab2022conditional, wang2024sea++}, meteorology \cite{bi2023accurate, wu2023interpretable, sun2022accurate}, and other domains \cite{sun2021solar, liu2023cross, sun2025hcan}. 
The key to most time series tasks lies in effectively capturing the temporal dynamics and underlying patterns that govern the evolution of data over time \cite{ma2024survey, wang2026deep}.
In recent years, deep learning has demonstrated remarkable success in uncovering complex dependencies and nonlinear structures within time series, giving rise to a variety of architectures such as multilayer perceptron (MLP) \cite{zeng2023dlinear}, convolutional neural network (CNN) \cite{eldele2024tslanet}, graph neural network (GNN) \cite{jin2025towards} and Transformer-based models \cite{nie2023patchtst}.
However, as illustrated in Figure~\ref{fig:Unified-MoE}~(a), existing approaches typically assume a single homogeneous distribution across all samples, overlooking the intrinsic pattern heterogeneity and regime shifts that frequently occur in real-world temporal data \cite{liu2024itransformer, nie2023patchtst, zeng2023dlinear}.
To address these challenges, pattern-aware time series modeling has emerged as a promising direction, aiming to explicitly recognize and model diverse temporal or frequency patterns for more robust and interpretable forecasting \cite{shao2024exploring}.
Among these advances, Mixture-of-Experts (MoE) methods have recently gained increasing attention for their ability to achieve both modeling diversity and computational efficiency \cite{wu2025enhancing}, making them a compelling paradigm for large-scale, pattern-aware time series analysis, as shown in Figure~\ref{fig:Unified-MoE}~(b).

The MoE framework enables a model to adaptively decompose complex data distributions into multiple specialized subspaces, which naturally aligns with the goal of pattern-aware modeling \cite{sun2024tfps}.
By integrating a set of specialized experts and a gating mechanism, MoE achieves both modeling diversity and computational efficiency \cite{liu2025moiraimoe}.
Each expert learns to capture a specific subspace or pattern within the data, while the gating network dynamically routes each input to the most relevant subset of experts.
Through this sparse activation strategy, MoE enables a large model capacity without a proportional increase in computational cost, allowing the system to scale efficiently on massive datasets and complex tasks \cite{xiaoming2025timemoe}.
With the help of this adaptive and selective routing mechanism, MoE has recently achieved remarkable success in the domains of Computer Vision (CV) \cite{zhang2023robust, cao2023MoE-Fusion} and Natural Language Processing (NLP) \cite{liu2024deepseekv2, dai2024deepseekmoe, teamqwen2024qwenmoe}, demonstrating its ability to model heterogeneous patterns and enhance scalability in large-scale learning systems.

With the great success of MoE in CV and NLP, it is appealing to extend MoE to time series data.
However, transferring the design for CV/NLP directly to time series data is non-trivial, and often fails to work in many scenarios \cite{feghali2021overview}.
Here, we highlight some typical challenges that arise when applying MoE to time series data.
First, the gating mechanism in MoE must capture temporal continuity and evolving dependencies to adaptively assign experts as distributions shift over time.
In fact, most MoE architectures designed for CV and NLP make static, token-level routing decisions under the assumption of independent inputs, which cannot effectively model the continuous and dynamic nature of time series data.
Second, experts in MoE need to model complex temporal dependencies within their subspaces and jointly capture heterogeneous temporal patterns across variables, scales, and regimes \cite{sun2025talon}.
This is different from experts in CV and NLP that are primarily designed for spatial or semantic diversity, making them ill-suited for modeling the temporal and multi-scale heterogeneity inherent in time series data.

\begin{figure}[t]
    \centering
    \includegraphics[width=0.9\linewidth]{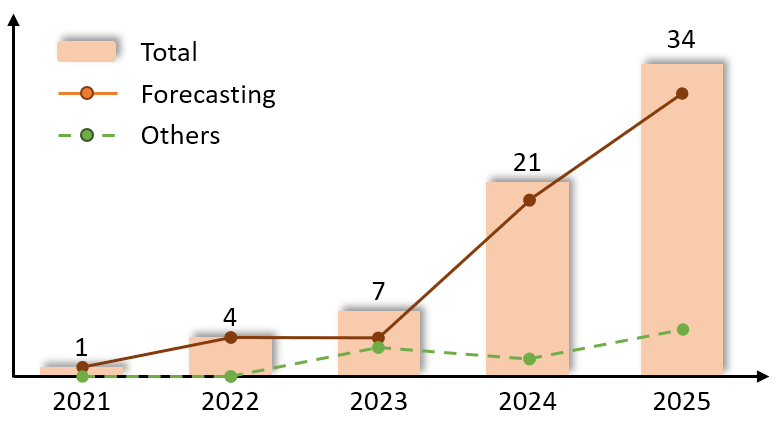}
    \caption{Number of publications on Mixture-of-Experts for time series modeling (2021–2025).}
    \label{fig:Trend}
\end{figure}

To the best of our knowledge, there has been no comprehensive or systematic surveys dedicated to MoE for time series data, despite the body of literature on MoE for time series \cite{xiaoming2025timemoe, liu2025moiraimoe, sun2024tfps}. 
As shown in Figure \ref{fig:Trend}, the number of publications on time series MoE has grown rapidly in recent years, especially in forecasting tasks, underscoring the increasing research attention and the need for a systematic review.
Although a few surveys share partial similarities with our work \cite{mu2025comprehensive}, they only focus on some representative MoE models for time series \cite{xiaoming2025timemoe}, and therefore fail to capture the broader methodological landscape.
Moreover, current reviews offer little discussion on open challenges and potential research directions, highlighting the need for a more systematic and forward-looking synthesis of time series MoE research.

This paper provides a comprehensive review of pattern-aware time series modeling, with a particular emphasis on the rapid emergence of MoE–based approaches. 
We begin by summarizing existing surveys on MoE and time series analysis, and then propose a new taxonomy that categorizes existing methods from two complementary perspectives: \emph{Gating Strategy} and \emph{Expert Design}. 
Specifically, the gating strategy is grouped into three types: ungated, dense gate, and sparse gate.
The expert design is divided into three structural forms: homogeneous, heterogeneous, and shared experts. 
The overall taxonomy is illustrated in Figure~\ref{fig:Taxonomy}. 
Finally, we conclude by outlining key open challenges and future research directions for time-series MoE, including channel-heterogeneous pattern modeling, automated expert discovery, and improved efficiency and scalability in structural design; self-evolving, reasoning-centric, agentic MoE for adaptive intelligence; and privacy-preserving mechanisms for trustworthy deployment.

Our main contributions are summarized as follows.
\begin{itemize}
    \item \textbf{New taxonomy and comprehensive review.}
    We present the first systematic and up-to-date survey of MoE methods for time series, covering studies from the past five years. 
    Existing works are organized into six coherent categories based on core architectural principles, offering a unified perspective that clarifies how time-series MoE has evolved across different modeling paradigms.
    \item \textbf{Cross-level insight into MoE mechanisms.}
    Through a dual-perspective analysis of routing mechanisms and expert architectures, we reveal how gating and experts interact to achieve temporal specialization, multi-scale coordination, and cross-variable adaptation.
    This analysis also uncovers shared design principles across seemingly different MoE models and explains why certain architectures are particularly robust to temporal distribution shifts.
    \item \textbf{Abundant future direction.} 
    We outline emerging research opportunities from three complementary dimensions: 
    Structural Design, focusing on channel-wise relational modeling, automated expert organization, and scalable architectures; 
    Adaptive Intelligence, emphasizing continual adaptation, temporal reasoning, and agentic expert behaviors; 
    Trustworthy Deployment, highlighting privacy preservation and robust deployment in multi-entity time-series settings. 
    These insights aim to guide the next generation of MoE research in time-series modeling.
\end{itemize}

\tikzstyle{my-box}=[
    rectangle,
    draw=hidden-draw,
    rounded corners,
    text opacity=1,
    minimum height=1.5em,
    minimum width=5em,
    inner sep=2pt,
    align=center,
    fill opacity=.5,
    line width=0.8pt,
]
\tikzstyle{leaf}=[my-box, minimum height=1.5em,
    fill=hidden-pink!80, text=black, align=left, 
    font=\normalsize,
    inner xsep=2pt,
    inner ysep=4pt,
    line width=0.8pt,
]
\begin{figure*}[!t]
    \centering
    \resizebox{0.9\textwidth}{!}
    {
        \begin{forest}
            forked edges,
            for tree={
                fill=level0!80,
                grow=east,
                reversed=true,
                anchor=base west,
                parent anchor=east,
                child anchor=west,
                base=left,
                font=\large,
                rectangle,
                draw=hidden-draw,
                rounded corners,
                align=left,
                minimum width=4em,
                edge+={darkgray, line width=1pt},
                s sep=3pt,
                inner xsep=2pt,
                inner ysep=4pt,
                line width=1.0pt,
                ver/.style={rotate=90, child anchor=north, parent anchor=south, anchor=center},
            },
            where level=1{text width=7.5em,font=\normalsize,fill=level1!80,}{},
            where level=2{text width=11.5em,font=\normalsize,fill=level2!80,}{},
            where level=3{text width=9em,font=\normalsize,fill=level3!60,}{},
            where level=4{text width=8.5em,font=\normalsize,fill=level4!20,}{},
            where level=5{text width=5em,font=\normalsize,}{},
            [\textbf{Mixture of Experts (MoE) for Time Series}, ver, minimum width=12em, minimum height=2.5em, inner xsep=6pt, inner ysep=6pt
                [Gating Strategy \\ (Sec.~\ref{sec:gating strategy})
                    [Ungated (Sec.~\ref{sec:ungated})
                        [Film~\cite{zhou2022film}{,} 
                        COST~\cite{woo2022cost}{,}
                        PTN~\cite{yang2025PTN}{,}
                        LeMoLE~\cite{zhang2024lemole}{,}
                        VPF-MoE~\cite{zhao2025vpfmoe}, leaf, text width=42.5em
                        ]
                    ]
                    [Dense Gate (Sec.~\ref{sec:densegate})
                        [Linear
                            [MoE-KAN~\cite{peng2024moekan}{,}
                            STGCN-MoE~\cite{chattopadhyay2022mixture}{,}
                            MoE-Traffic~\cite{ke2024interpretable}{,}
                            TSEPMoE~\cite{chen2023tsepmoe}{,} \\
                            IMMOE~\cite{tong2023immoe}{,}
                            DeepUnifiedMoM~\cite{ong2024deepunifiedmom}{,}
                            CAD~\cite{si2023cad}{,}
                            FEDformer~\cite{zhou2022fedformer}{,} \\
                            TCNet~\cite{zhang2023TCNet}{,} 
                            MambaMoE~\cite{wang2025mambamoe}{,}
                            AME~\cite{zhang2024ame}{,}
                            DFMH~\cite{dong2025dfmh}{,}
                            SMETimes~\cite{fan2025smetimes}{,} \\
                            MoPE~\cite{niu2024mope}{,}
                            VE~\cite{wang2024ve}{,}
                            FreqMoE~\cite{liu2025freqmoe}{,}  
                            WaveTS-M~\cite{zhou2026wavelet}, leaf, text width=31.8em]
                        ]
                        [Attention
                            [MAES~\cite{bourgin2021maes}{,} DynaMix~\cite{hemmer2026true}{,} MoME~\cite{li2025mome}{,} Graph-MoE~\cite{huang2025graphmoe}{,} TITAN~\cite{wang2024titan}{,} \\ CICADA~\cite{lan2025cicada}, leaf, text width=31.8em]
                        ]
                        [LSTM
                            [Dychem~\cite{han2022dychem}{,} TMMOE~\cite{yuan2023tmmoe}, leaf, text width=31.8em]
                        ]
                        [Non-parametric
                            [MSGNet~\cite{cai2024msgnet}{,} MoE-F~\cite{saqur2025moef}{,} MOOE~\cite{zhao2025mooe}, leaf, text width=31.8em]
                        ]
                    ]
                    [Sparse Gate (Sec.~\ref{sec:sparsegate})
                        [Parametric
                            [Without balancing
                                [MoE-AMC~\cite{gao2023moeamc}{,} 
                                ARTEMIS~\cite{feng2025artemis}{,} xTime~\cite{li2025xtime}{,} \\ MGTS-Net~\cite{cai2024msgnet}, leaf, text width=21.8em]
                            ]
                            [Noise injection
                                [Pathformer~\cite{chen2024pathformer}{,} 
                                Super-Linear~\cite{nochumsohn2025superlinear}{,} \\
                                TransMoE~\cite{shi2025transmoe}{,}
                                MoU~\cite{peng2025mou}, leaf, text width=21.8em]
                            ]
                            [Balancing loss
                                [SoftShape~\cite{liu2025softshape}{,}
                                BP-MOE~\cite{chen2024bpmoe}{,}
                                TopK Dense~\cite{sievers2024softdense}{,} \\
                                Time-MoE~\cite{xiaoming2025timemoe}{,}
                                ContexTST~\cite{hong2025contextst}{,} \\
                                MoFE-Time~\cite{liu2025mofetime}{,}
                                CAPTime~\cite{yao2025captime}, leaf, text width=21.8em]
                            ]
                            [Noise and loss
                                [MMK~\cite{han2024MMK}
                                TALON~\cite{sun2025talon}{,}
                                MixMamba~\cite{alkilane2024mixmamba}{,} \\
                                TimrFilter~\cite{hu2025timefilter}, leaf, text width=21.8em]
                            ]
                        ]
                        [Similarity-driven
                            [TFPS~\cite{sun2024tfps}{,}
                            Moirai-MoE~\cite{liu2025moiraimoe}{,}
                            Time Tracker~\cite{shi2025timetracker}{,} 
                            ProtoN-FM~\cite{gong2026bridging}{,} \\
                            FuseMoE~\cite{han2024fusemoe}{,} 
                            FT-MoE~\cite{xiao2025ftmoe}{,}
                            TimeExpert~\cite{ma2025timeexpert}, leaf, text width=31.8em]
                        ]
                    ]
                ]
                [Expert Design \\ (Sec.~\ref{sec:expert design})
                    [Homogeneous Experts \\ (Sec.~\ref{sec:homoexperts})
                        [FFN
                            [Small models
                                [MoPE~\cite{niu2024mope}{,} VE~\cite{wang2024ve}{,} FuseMoE~\cite{han2024fusemoe}{,} \\ FreqMoE~\cite{liu2025freqmoe}{,} AMD~\cite{hu2025amd}{,} MoME~\cite{li2025mome}{,} \\ TFPS~\cite{sun2024tfps}{,} MOOE~\cite{zhao2025mooe}{,} WaveTS-M~\cite{zhou2026wavelet}{,} \\ MoU~\cite{peng2025mou}{,} FT-MoE~\cite{xiao2025ftmoe}{,} TimeExpert~\cite{ma2025timeexpert}{,} \\ MGTS-Net~\cite{hao2025mgts}{,} LeMoLE~\cite{zhang2024lemole}{,} \\ Soft/TopK Dense~\cite{sievers2024softdense}{,} TransMoE~\cite{shi2025transmoe}{,} \cite{yumlu2003financial}{,} \\ ANFIS~\cite{ebrahimpour2011anfis}, leaf, text width=21.5em]
                            ]
                            [Large models
                                [SMETimes~\cite{fan2025smetimes}{,} CAPTime~\cite{yao2025captime}{,} \\ Moirai-MoE~\cite{liu2025moiraimoe}{,} TIME-MoE~\cite{xiaoming2025timemoe}{,} \\ ULoRA-MoE~\cite{wuuncertaintyuloramoe}, leaf, text width=21.5em]
                            ]
                        ]
                        [LSTM
                            [Soft/TopK LSTM~\cite{sievers2024softdense}{,} TMMOE~\cite{yuan2023tmmoe}{,} DeepUnifiedMoM~\cite{ong2024deepunifiedmom}, leaf, text width=31.8em]
                        ]
                        [Attention
                            [Pathformer~\cite{chen2024pathformer}{,} TITAN~\cite{wang2024titan}{,} DFMH~\cite{dong2025dfmh}{,} POND~\cite{wang2024pond}{,} Graph-MoE~\cite{huang2025graphmoe}, leaf, text width=31.8em ]
                        ]
                        [Mamba
                            [MixMamba~\cite{alkilane2024mixmamba}{,} MambaMoE~\cite{wang2025mambamoe}, leaf, text width=31.8em ]
                        ]
                        [Others
                            [STGCN-MoE~\cite{chattopadhyay2022mixture}{,} Msgnet~\cite{cai2024msgnet}{,} TimeFilter~\cite{hu2025timefilter}{,} CoST~\cite{woo2022cost}{,} TCNet~\cite{zhang2023TCNet}{,} \\ TFMoE~\cite{lee2024tfmoe}{,} MMK~\cite{han2024MMK}, leaf, text width=31.8em ]
                        ]
                        [Custom
                            [CAD~\cite{si2023cad}{,} MoE-traffic~\cite{ke2024interpretable}{,} xTime~\cite{li2025xtime}{,} DynaMix~\cite{hemmer2026true}{,} MoFE-Time~\cite{liu2025mofetime}{,} \\ PTN~\cite{yang2025PTN}{,} ProtoN-FM~\cite{gong2026bridging}{,} FEDformer~\cite{zhou2022fedformer}{,} Film~\cite{zhou2022film}{,} MoE-KAN~\cite{peng2024moekan}, leaf, text width=31.8em ]
                        ]
                    ]
                    [Heterogeneous Experts \\ (Sec.~\ref{sec:heteroexperts})
                        [Statistical - neural
                            [Dychem~\cite{han2022dychem}{,} TSEPMoE~\cite{chen2023tsepmoe}, leaf, text width=31.8em
                            ]
                        ]
                        [Neural models
                            [TALON~\cite{sun2025talon}{,}  MoME~\cite{li2025mome}{,} MGTS-Net~\cite{hao2025mgts}{,} IMMOE~\cite{tong2023immoe}{,} \cite{garg2024time}{,} \cite{vallarino2024dynamic}{,} \\ MoE-AMC~\cite{gao2023moeamc}{,} MAES~\cite{bourgin2021maes}{,} BP-MoE~\cite{chen2024bpmoe}, leaf, text width=31.8em
                            ]
                        ]
                        [Foundation models
                            [MoE-F~\cite{saqur2025moef}{,} VPF-MoE~\cite{zhao2025vpfmoe}{,} MSE-ITT~\cite{koval2025MSEITT}, leaf, text width=31.8em
                            ]
                        ]
                    ]
                    [Shared Experts (Sec.~\ref{sec:sharedexperts})
                        [MoLE~\cite{ni2024mole}{,} SoftShape~\cite{liu2025softshape}{,} AME~\cite{zhang2024ame}{,} Time-MoE~\cite{xiaoming2025timemoe}{,} Time Tracker~\cite{shi2025timetracker}{,} ContexTST~\cite{hong2025contextst}{,} \\ PatchMoE~\cite{wu2025PatchMoE}{,} ARTEMIS~\cite{feng2025artemis}, leaf, text width=42.5em ]
                    ]
                ]
                [Outlook \\ (Sec.~\ref{sec:future})
                    [Structural Design \\ (Sec.~\ref{sec:structural})
                        [Channel-wise relational modeling{,} Automated temporal expert discovery{,} Efficient \& scalable \\ TS MoE, leaf, text width=42.5em
                        ]
                    ]
                    [Adaptive Intelligence \\ (Sec.~\ref{sec:adaptive})
                        [Self-evolving MoE for non-stationarity{,} Temporal reasoning-centric MoE{,} Agentic MoE for \\ TS intelligence, leaf, text width=42.5em
                        ]
                    ]
                    [Trustworthy Deployment \\ (Sec.~\ref{sec:trustworthy})
                        [Privacy-preserving multi-entity TS MoE, leaf, text width=42.5em
                        ]
                    ]
                ]
            ]
        \end{forest}
    }
\vspace{-0.5em}
\caption{A comprehensive taxonomy of MoE for time series modeling, categorized according to routing strategy and expert design.}
\vspace{-1em}
\label{fig:Taxonomy}
\end{figure*}
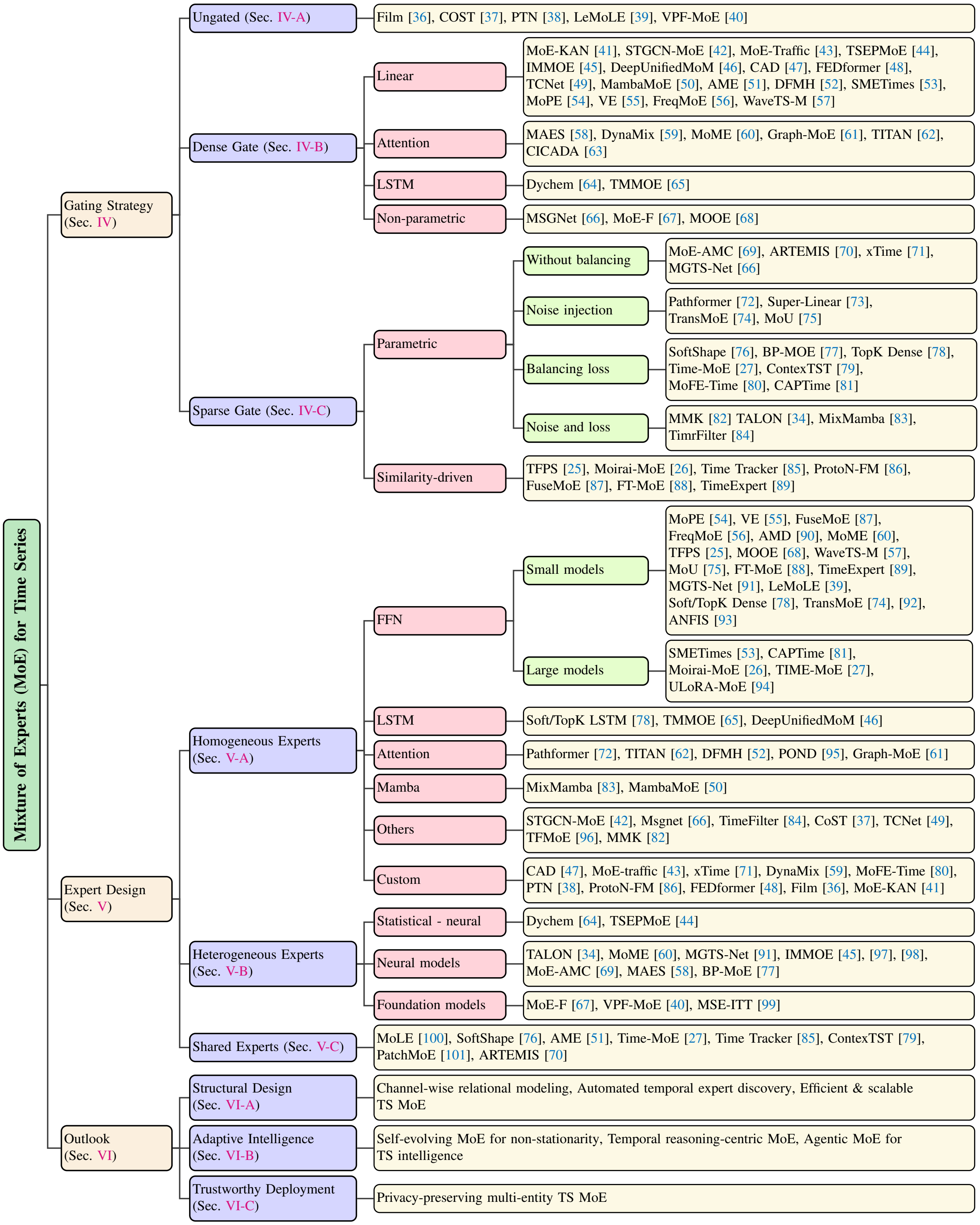

\section{Related surveys}
\label{sec:related}
\subsection{Overview of Time Series Modeling}
Time series data refer to ordered sequences of observations indexed over time, where each value depends on its historical context rather than being independently and identically distributed \cite{li2023transferable,zamanzadeh2024deep,ning2026ts}. 
This temporal dependency manifests across multiple scales, from local autocorrelations to long-range dependencies, and fundamentally distinguishes time series from other data modalities.
A univariate time series can be expressed as $\mathbf{X} = (x_1, x_2, \dots, x_T)$, 
where $x_t \in \mathbb{R}$ denotes the observation at timestamp $t$, 
while a multivariate time series extends this to $\mathbf{X} = [\mathbf{x}_1, \mathbf{x}_2, \dots, \mathbf{x}_T]$, 
where each $\mathbf{x}_t \in \mathbb{R}^C$ represents a $C$-dimensional measurement at time $t$ \cite{nie2023patchtst, wang2024sea++}.

Real-world time series exhibit remarkable heterogeneity \cite{shao2024exploring, cai2025time}. 
Physiological signals follow periodic biological rhythms, financial markets fluctuate with regime shifts, and industrial sensors generate irregular and noisy traces. 
Moreover, factors such as non-stationarity, distribution drift, missing values, and uneven sampling further complicate modeling \cite{luo2024knowledge}. 
These challenges yield diverse temporal behaviors, ranging from smooth seasonal cycles to volatility bursts and abrupt anomalies.

Unlike images or text, which have standardized representations (pixel grids, token vocabularies), time series values are continuous, domain-dependent, and context-sensitive \cite{sun2025talon}. 
They lack a unified encoding space transferable across domains. 
This absence of standardized representations makes general-purpose modeling particularly challenging. 

The diversity and irregularity of time series suggest that a single model architecture cannot effectively capture all temporal structures. 
Instead, partitioning data into distinct regimes and assigning specialized components to different dynamics becomes essential. 
This motivates the use of \textit{Mixture-of-Experts} (MoE) architectures, where multiple experts specialize in modeling different temporal patterns, including seasonal versus irregular behaviors, short-term versus long-term dynamics, and stationary versus evolving regimes, thereby providing a natural framework for robust and adaptive time-series modeling.

\begin{table}[!t]
\caption{An overview of recent MoE surveys on different modalities. The yellow part indicates the proportion of this modality in all five modalities. It can be seen that this paper is mainly concerned with MoE on time series data.}
\label{tab:surveys}
\centering
\begin{tabular}{c|c|cccccc} \toprule	
\multirow{2}{*}{Paper}  & \multirow{2}{*}{Year} & \multicolumn{5}{c}{Modality} \\  \cmidrule(lr){3-7}

 & & Text & Image & Video & Audio & Time Series \\ \midrule

\cite{fedus2022review} & 2022 &
$\includegraphics[width=5mm]{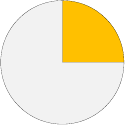}$ & 
$\includegraphics[width=5mm]{Figure/mod_14.pdf}$&
$\includegraphics[width=5mm]{Figure/mod_14.pdf}$&
$\includegraphics[width=5mm]{Figure/mod_14.pdf}$ &
$\includegraphics[width=5mm]{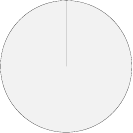}$ \\ \midrule 

\cite{dimitri2025survey} & 2025 &
$\includegraphics[width=5mm]{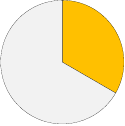}$ &
$\includegraphics[width=5mm]{Figure/mod_13.pdf}$&
$\includegraphics[width=5mm]{Figure/mod_0.pdf}$&
$\includegraphics[width=5mm]{Figure/mod_13.pdf}$&
$\includegraphics[width=5mm]{Figure/mod_0.pdf}$ \\ \midrule 

\cite{al2025moe}& 2025 &
$\includegraphics[width=5mm]{Figure/mod_13.pdf}$ &
$\includegraphics[width=5mm]{Figure/mod_13.pdf}$&
$\includegraphics[width=5mm]{Figure/mod_0.pdf}$&
$\includegraphics[width=5mm]{Figure/mod_13.pdf}$&
$\includegraphics[width=5mm]{Figure/mod_0.pdf}$ \\ \midrule 

\cite{cai2025survey} & 2025 &
$\includegraphics[width=5mm]{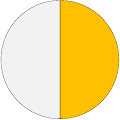}$ &
$\includegraphics[width=5mm]{Figure/mod_12.pdf}$&
$\includegraphics[width=5mm]{Figure/mod_0.pdf}$&
$\includegraphics[width=5mm]{Figure/mod_0.pdf}$&
$\includegraphics[width=5mm]{Figure/mod_0.pdf}$ \\ \midrule 

\cite{zhang2025mixture} & 2025 &
$\includegraphics[width=5mm]{Figure/mod_12.pdf}$ &
$\includegraphics[width=5mm]{Figure/mod_12.pdf}$&
$\includegraphics[width=5mm]{Figure/mod_0.pdf}$&
$\includegraphics[width=5mm]{Figure/mod_0.pdf}$&
$\includegraphics[width=5mm]{Figure/mod_0.pdf}$ \\ \midrule 

\cite{mu2025comprehensive} & 2025 &
$\includegraphics[width=5mm]{Figure/mod_14.pdf}$ &
$\includegraphics[width=5mm]{Figure/mod_14.pdf}$&
$\includegraphics[width=5mm]{Figure/mod_14.pdf}$&
$\includegraphics[width=5mm]{Figure/mod_0.pdf}$&
$\includegraphics[width=5mm]{Figure/mod_14.pdf}$ \\ \midrule 

\cite{liu2026survey} & 2026 &
$\includegraphics[width=5mm]{Figure/mod_13.pdf}$ &
$\includegraphics[width=5mm]{Figure/mod_13.pdf}$&
$\includegraphics[width=5mm]{Figure/mod_0.pdf}$&
$\includegraphics[width=5mm]{Figure/mod_13.pdf}$&
$\includegraphics[width=5mm]{Figure/mod_0.pdf}$ \\ \midrule 

\textbf{Ours} & 2026 & 
$\includegraphics[width=5mm]{Figure/mod_0.pdf}$&
$\includegraphics[width=5mm]{Figure/mod_0.pdf}$&
$\includegraphics[width=5mm]{Figure/mod_0.pdf}$&
$\includegraphics[width=5mm]{Figure/mod_0.pdf}$&
$\includegraphics[width=5mm]{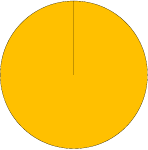}$ \\
\bottomrule
\end{tabular}
\end{table}

\subsection{Surveys on Mixture-of-Experts}
In recent years, several surveys have systematically reviewed the development of MoE models from diverse perspectives, including architectural evolution, optimization, large-scale deployment, and domain-specific applications. 
Table~\ref{tab:surveys} summarizes representative MoE surveys and their modality coverage, allowing readers to quickly locate works relevant to their research interests.

Most existing surveys primarily focus on text, image, and multimodal domains, aiming to characterize MoE progress in large language models (LLMs) or large-scale Transformer systems.
The earliest effort on sparse expert models \cite{fedus2022review} discusses routing sparsity, load balancing, and stability.
Later surveys \cite{liu2026survey, dimitri2025survey, al2025moe, mu2025comprehensive, cai2025survey, zhang2025mixture} extend the discussion toward historical evolution, inference optimization, system deployment, theoretical unification, and LLM-specific adaptation, respectively.
These works collectively outline how MoE research has evolved from efficient routing to modular foundation architectures.

Despite these advances, existing MoE surveys remain largely fragmented and modality-biased. 
They can be roughly categorized along two orthogonal perspectives: \emph{(i) gating paradigms}, which define how experts are selected or activated; and \emph{(ii) expert architectures}, which determine the representational capacity and specialization.
These dimensions also form the organizational backbone of our taxonomy in Section~\ref{sec:gating strategy} and~\ref{sec:expert design}.

Compared with prior surveys, our work represents, to the best of our knowledge, the first comprehensive review dedicated to MoE methods for time-series modeling.
We bridge statistical forecasting, neural MoE paradigms, and foundation-scale temporal modeling, providing both a unified taxonomy and forward-looking perspectives.

\subsection{Surveys on Time Series Data}
Existing surveys on time series modeling can be broadly categorized into three groups according to their research focus.

\textbf{(1) Task-oriented surveys.}
These works summarize methods developed for specific downstream tasks, such as forecasting \cite{lim2021forecasting, shao2024forecasting}, classification \cite{ismail2019classification}, imputation \cite{wang2025deep}, and anomaly detection \cite{zamanzadeh2024anomaly}. 
They mainly compare learning objectives, model architectures, and evaluation protocols across different applications, providing a comprehensive view of how task-specific challenges shape the evolution of time series algorithms.

\textbf{(2) Method-oriented surveys.}
A second line of reviews categorizes time series research from the perspective of model architectures and learning paradigms. 
Representative works \cite{wen2023transformers, jin2024graph, yang2026survey, liang2024foundation} analyze transformer-based forecasting models, state-space approaches such as Mamba, and the emerging trend of time series foundation models. 
These surveys trace the architectural evolution from sequence modeling to large-scale and modularized designs, emphasizing scalability, representation sharing, and cross-domain generalization. 
This methodological progression reflects an increasing demand for flexible architectures capable of adapting to diverse temporal patterns and heterogeneous data distributions.

\textbf{(3) Technology-oriented surveys.}
More recent efforts investigate broader methodological trends, including self-supervised learning \cite{eldele2023self, zhang2024ssl}, continual learning \cite{besnard2024continual}, and multimodal representation learning \cite{liu2025multimodalities}, and privacy-preserving modeling \cite{mao2024differential}. 
Although these topics extend beyond conventional model design, they share a common pursuit of adaptability, efficiency, and specialization, which also underpin modern expert-based frameworks.

Taken together, these surveys illustrate the gradual evolution of time series research from task-driven analysis to architecture-driven and technology-integrated paradigms. 
As temporal modeling increasingly emphasizes modularity, adaptivity, and scalability, the MoE paradigm naturally emerges as a unifying framework that embodies these principles. 
Our survey therefore consolidates MoE research within the context of time series modeling, offering a structured taxonomy and theoretical foundation that connects expert-based architectures with distributional heterogeneity and dynamic specialization.

\section{Formalization of Mixture-of-Experts}
\label{sec:Formalization of MoE}
The concept of the Mixture-of-Experts (MoE) model has a long history \cite{jacobs1991adaptive, jordan1994hierarchical}, and recent advances in large LLMs have further revived its use in large-scale learning systems \cite{kim2024scaling, xue2024openmoe}. 
MoE provides a flexible modeling paradigm designed to handle data heterogeneity by combining multiple specialized predictors. 
The key idea is to partition the input space into regions where different experts perform best and to dynamically route each input to the most suitable expert(s).

Formally, given a time series $\mathbf{X} = [\mathbf{x}_1, \mathbf{x}_2, \dots, \mathbf{x}_T]$, 
an MoE framework is composed of \emph{expert networks} and a \emph{gating network}, 
where the latter controls both expert activation and the aggregation of their outputs. 
For simplicity, the following formalization focuses on the univariate case ($C=1$); 
the same principles naturally extend to multivariate settings. 
In practice, each token $\mathbf{x}_t$ can represent either a \textit{point-level} observation (a single time step) \cite{xiaoming2025timemoe} or a \textit{patch-level} feature obtained by segmenting consecutive points \cite{sun2024tfps}, depending on the temporal granularity of the MoE design.

\subsection{Expert Networks}
A collection of $M$ expert functions is defined as:
\begin{align}
    f_m(\mathbf{x}_t; \theta_m), \quad m = 1, \dots, M,
\end{align}
where $\theta_m$ denotes the parameters of the $m$-th expert. 
Each expert learns a sub-function of the overall mapping, focusing on particular patterns or regions of the input space.

\subsection{Gating Strategies}
The gating network determines how each token $\mathbf{x}_t$ is assigned to one or more experts. 
Two primary gating strategies are widely adopted in MoE frameworks: dense and sparse gating.

\textbf{Dense Gate.}
The dense gating function $\mathcal{G}(\mathbf{x}_t; \phi)$ produces a probability distribution over experts:
\begin{align}
    \mathcal{G}(\mathbf{x}_t; \phi) = \operatorname{softmax}(g(\mathbf{x}_t; \phi)),
\end{align}
where $\phi$ denotes the parameters of the gating network, and $g(\mathbf{x}_t; \phi)$ represents the pre-softmax gating logits.
Under dense gating, all experts participate in prediction, and their outputs are combined via weighted summation:
\begin{align}
\label{eq:dense_output}
    y_t = \sum_{m=1}^{M} \mathcal{G}(\mathbf{x}_t; \phi)_m \, f_m(\mathbf{x}_t; \theta_m).
\end{align}
This structure, illustrated in Figure~\ref{fig:Routing}~(b), ensures smooth gradient flow across all experts but incurs high computational overhead since every expert is activated for each input.

\textbf{Sparse Gate.}
Although dense MoE architectures can achieve high accuracy, they are computationally expensive because all experts are active for every token. 
To address this, \cite{shazeer2017outrageously} proposed the sparsely gated MoE, which activates only the top-$k$ experts with the highest gating scores during each forward pass.
The sparse gating mechanism is expressed as:
\begin{align}
\label{eq:sparse-gate}
\mathcal{G}(\mathbf{x}_t; \phi)_m =
\operatorname{softmax}\!\left(
\operatorname{TopK}\!\left(g(\mathbf{x}_t; \phi), k\right)
\right)_m,
\end{align}
\begin{align}
\operatorname{TopK}(g(\mathbf{x}_t; \phi), k)_m =
\begin{cases}
g(\mathbf{x}_t; \phi)_m, & \text{if condition}, \\
-\infty, & \text{otherwise},
\end{cases} \nonumber
\end{align}
\begin{align}
\text{condition: } 
g(\mathbf{x}_t; \phi)_m \text{ is among the top-}k \text{ entries of } g(\mathbf{x}_t; \phi).
\end{align}
The $\operatorname{TopK}(\cdot, k)$ operator preserves the $k$ largest gating values and masks the rest with $-\infty$, ensuring that the subsequent softmax assigns near-zero activation to inactive experts.

The resulting sparse MoE output is:
\begin{align}
\label{eq:sparse_output}
    y_t = \sum_{m=1}^{M} \mathcal{G}(\mathbf{x}_t; \phi)_m\, f_m(\mathbf{x}_t; \theta_m),
\end{align}
where only the top-$k$ components of $\mathcal{G}(\mathbf{x}_t; \phi)$ are nonzero.

\subsection{Why MoE Fits Time Series Modeling}
\label{sec:MoE for TSM}

Although the MoE framework was originally proposed as a general modeling paradigm, its structural properties align naturally with the intrinsic challenges of time series data.
Temporal signals commonly exhibit heterogeneous regimes, including periodic patterns, trend shifts, and irregular fluctuations, which are difficult for a single, monolithic model to capture \cite{sun2024tfps}. The expert specialization mechanism of MoE enables different modules to model distinct temporal behaviors, while the gating network adaptively routes each input segment to the most relevant experts based on its local context. In addition, sparse MoE architectures offer scalability by activating only a small subset of experts per token, allowing large expert pools to be deployed without excessive computational cost.
These characteristics have driven increasing adoption of MoE in time series forecasting, anomaly detection, and representation learning.
Sections~\ref{sec:gating strategy} and~\ref{sec:expert design} discuss gating strategies and expert architectures designed specifically for time series modeling.

\section{Gating Strategy}
\label{sec:gating strategy}

The gating strategy, also known as the gating or routing function, is a key component in MoE architectures. It controls the activation and aggregation of experts by assigning importance weights to each, thereby determining both expert participation and their contribution to the final prediction. Through this process, MoE models can dynamically allocate computational capacity and promote expert specialization.
Depending on how the gating function activates experts and combines their outputs, time-series MoE models can be categorized into three groups: (1) \textbf{Ungated models}, which integrate expert outputs without an explicit gating function; (2) \textbf{Dense-gated models}, where all experts are activated and combined through continuous weighting; and (3) \textbf{Sparse-gated models}, which selectively activate a subset of experts based on routing decisions.

\begin{figure}[t]
\centering
\includegraphics[width=1.0\linewidth]{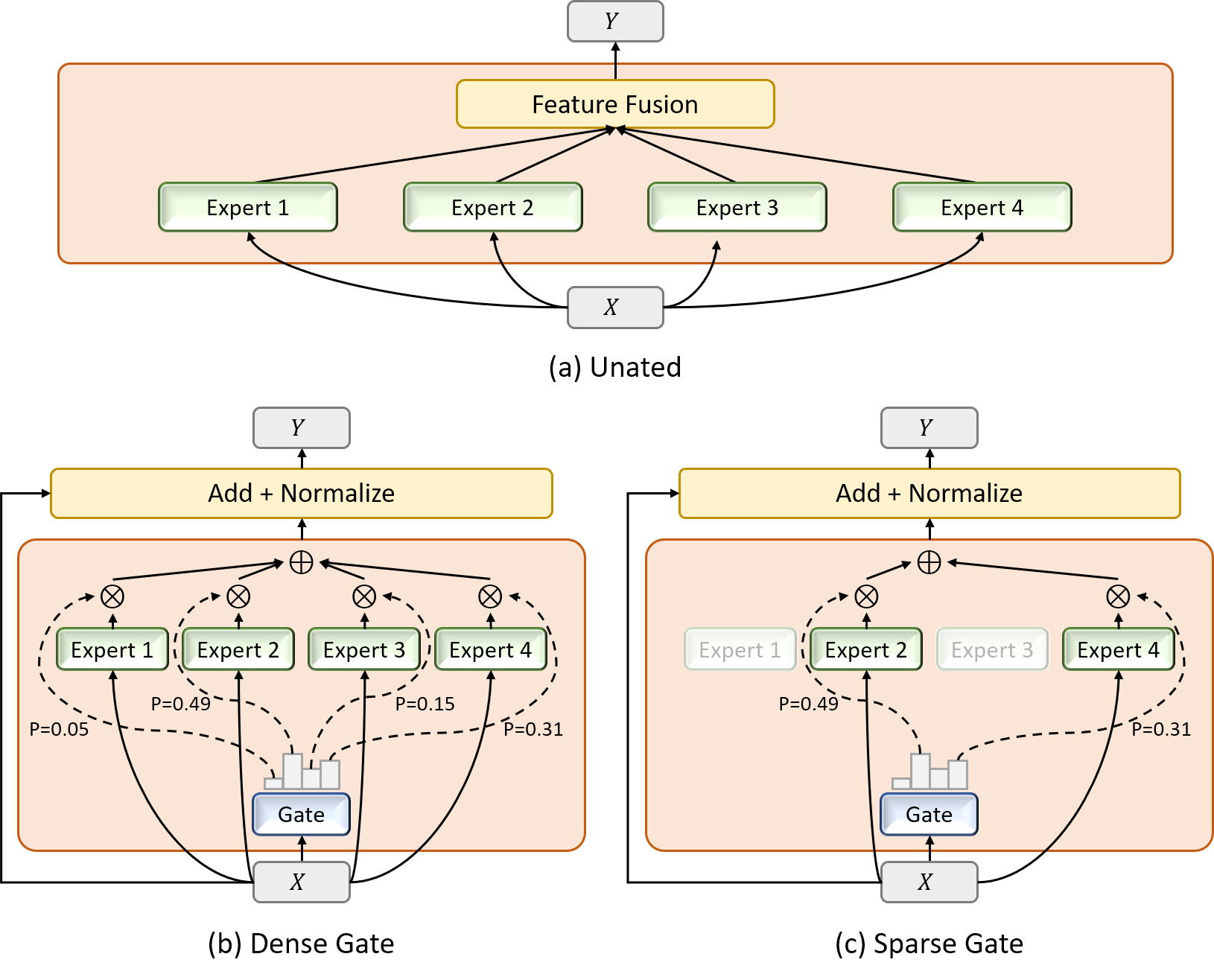}
\caption{Illustration of gating mechanisms in Mixture-of-Experts (MoE) models. Given an input $X$, the gating network produces expert logits that determine expert activation. In the ungated setting (a), the outputs of all experts are combined through a predefined fusion function without any routing. In the dense setting (b), all experts are activated and their outputs are weighted by the corresponding gating probabilities. In the sparse setting (c), only the top-$k$ experts are selected for conditional routing, and their outputs are scaled by their assigned gating weights.}
\label{fig:Routing}
\end{figure}

\subsection{Ungated}
\label{sec:ungated}
Ungated models preserve the multi-expert architecture of MoE frameworks but omit an 
explicit gating function that dynamically selects or weights experts, as shown in Figure~\ref{fig:Routing}~(a). Instead, all experts are simultaneously activated, and their outputs are fused through predefined or learnable aggregation mechanisms. 

For instance, FiLM \cite{zhou2022film} utilizes a mixture of multiscale experts and combines their outputs via a linear layer for long-term time series forecasting.
CoST \cite{woo2022cost} employs a mixture of auto-regressive experts and merges their representations through average pooling to disentangle trend components for time series representation learning.
PTN \cite{yang2025PTN} follows a sequential assignment strategy, where input patches are routed deterministically to experts without any gating or fusion among expert outputs.
LeMoLE \cite{zhang2024lemole} fuses multiple linear experts through CNN–based and FiLM-based \cite{perez2018film} modulation conditioned on frozen language model embeddings, achieving adaptive feature integration without explicit gating.
VPF-MoE \cite{zhao2025vpfmoe} similarly operates without a trainable router: it selects the best-performing expert model from a pool of candidates through an offline voting mechanism, and uses this static choice for prediction instead of relying on dynamic gating.

While this simplification improves training stability and reduces architectural complexity, it inherently limits the model’s ability to specialize experts or adapt computation to the input.

\subsection{Dense Gate}
\label{sec:densegate}

In dense-gated MoE models, all experts are simultaneously activated, and the gating function assigns continuous weighting coefficients to aggregate their outputs, as shown in Figure~\ref{fig:Routing}~(b). Through this process, every expert contributes to the overall computation, while the gating network learns to balance their relative importance.

Although the overall principle remains the same, the internal design of the gating network differs across architectures, resulting in distinct strategies for computing expert weights. According to their internal structure, dense-gated models can be broadly categorized into four types. 
(1) \textit{Linear-based Gate}, where expert weights are produced through a simple linear projection; (2) \textit{Attention-based Gate}, which derives expert weights from contextual relevance or similarity measures using attention mechanisms; (3) \textit{LSTM-based Gate}, which models temporal dependencies using recurrent structures such as Long Short-Term Memory (LSTM) before generating expert weights;  (4) \textit{Non-parametric gates}, which replace learnable neural gating networks with algorithmic or theoretically grounded mechanisms that compute expert weights without trainable parameters.

\subsubsection{Linear-based Gate} 
Linear-based gating constitutes one of the most fundamental and widely used forms of dense gating. 
\begin{align}
    g(\mathbf{x}_t; \phi) = W\mathbf{x}_t + b.
\end{align}
It computes expert weights through a simple linear projection followed by softmax normalization, making the mechanism conceptually straightforward and computationally efficient. Based on the type of input features used to generate the gating logits, linear-based gating methods can be further categorized into three subclasses.

\textit{(1)	Original time series as input.}
The gating mechanism directly operates on the original time-series, allowing expert weighting to be driven by raw temporal dynamics rather than compressed hidden embeddings. 
Specifically, MOE-KAN \cite{tong2023immoe}, STGCN-MoE \cite{chattopadhyay2022mixture}, and Traffic-MoE \cite{ke2024interpretable} employ a single dense gate on raw time-series inputs to weight multiple experts, while TSEPMoE \cite{chen2023tsepmoe} uses the same gating structure but takes exogenous variable series as inputs.
On the other hand, IMMOE \cite{tong2023immoe} employs multiple gates, where each gate takes the same raw series as input but generates different expert weightings for different forecast horizons.
DeepUnifiedMom \cite{ong2024deepunifiedmom} adopts a similar design for multi-scale momentum modeling with a hand-crafted time series momentum as input.
CAD \cite{si2023cad} further enhances this multi-gate approach by introducing shared and personalized gates, where the shared gate receives time series from all channels, while each personalized gate only operates on a specific channel. This input separation stabilizes gating and reduces cross-metric interference in expert weighting.

\textit{(2)	Time series representation as input.}
Using learned temporal representations as gate input enables expert routing to leverage informative temporal abstractions and higher-level dependencies, resulting in richer and more context-aware expert selection.
Some models represent this paradigm at the intermediate feature level, where they use linear gating over hidden representations within the feature extraction modules, but for different purposes. 
For instance, FEDformer \cite{zhou2022fedformer} and TCNet \cite{zhang2023TCNet} apply the gate to decomposition filters to separate temporal components at different scales, 
while MambaMoE \cite{wang2025mambamoe} and AME \cite{zhang2024ame} use gated experts to model temporal and channel dependencies, respectively. 
DFMH \cite{dong2025dfmh} and SMETimes \cite{fan2025smetimes} further extend this design by gating experts over enriched contextual embeddings. DFMH learns fusion weights over heterogeneous encoder representations, while SMETimes gates experts on LLM-enhanced segment embeddings.
MoPE \cite{niu2024mope} shifts gating to the output layer, where the backbone’s final latent representation is used to weigh multiple projection heads, enabling a more expressive alternative to a single shared linear predictor. 
VE \cite{wang2024ve} provides a special case of representation-based gating, where a learnable embedding lookup table serves as the gate input, allowing variate-specific routing.

\textit{(3)	Frequency-domain representation as input.}
Using frequency-domain representations as gate input allows the routing mechanism to exploit periodic structure and energy distribution across frequencies, providing an alternative spectral perspective for expert selection.
In WaveTS-M \cite{zhou2026wavelet}, the linear gate operates on low-frequency wavelet coefficients to weight experts according to their suitability for slow-varying components. FreqMoE \cite{liu2025freqmoe} generalizes this idea by gating experts based on the magnitude spectrum, allowing dynamic expert allocation across learned frequency bands.

\subsubsection{Attention-based Gate}
Attention-based dense gating extends traditional gating mechanisms by incorporating attention operations to compute expert weights from contextual relevance. 
\begin{align}
    g(\mathbf{x}_t; \phi)_m = \frac{q(\mathbf{x}_t)^T k_m}{\sqrt{d}}.
\end{align}
It dynamically evaluates the similarity between input representations and expert-specific embeddings or hidden states, enabling context-dependent weighting and adaptive fusion of expert outputs.

MAES \cite{bourgin2021maes} represents a canonical form of attention gating, where the gate computes similarity between a sequence-level context vector and expert embeddings, enabling adaptive expert fusion. 
DynaMix \cite{hemmer2026true} extends this idea by introducing state attention, where temporal context is first aggregated via attention before determining expert weights, yielding more context-sensitive routing.
MoME \cite{li2025mome} adopts a Transformer-based attention gate that models cross-domain dependencies and produces adaptive fusion weights for time-, frequency-, and spatial-domain experts.
Further, Graph-MoE \cite{huang2025graphmoe} and TITAN \cite{wang2024titan}introduce memory-augmented attention gating, allowing the gate to incorporate historical context and stabilize expert assignment over time. In Graph-MoE, the gate attends to a global memory of graph-informed states to assign soft expert weights across GNN layers, whereas in TITAN, the gate queries a memory representation alongside expert-specific representations, enabling memory-informed expert routing. 
Lastly, CICADA \cite{lan2025cicada} adopts a hierarchical attention gate that models both intra- and inter-expert interactions, generating a more structured and semantically grouped expert weighting pattern.

\subsubsection{LSTM-based Gate}
LSTM-based gating introduces explicit temporal dependency modeling into MoE architectures. 
Instead of computing expert weights from a single input token, these gates process sequential inputs through an LSTM to capture historical context, followed by a fully connected layer to produce time-varying mixing coefficients:
\begin{align}
    h_t &= \operatorname{LSTM}(\mathbf{x}_t, h_{t-1}), \\
    g(\mathbf{x}_t; \phi) &= W h_t + b.
\end{align}
For instance, DyCham \cite{han2022dychem} adopts an LSTM-based gate to fuse predictions from multiple forecasting models in hierarchical time-series settings. 
The LSTM processes windowed historical sequences, enabling the gate to adjust expert contributions according to evolving temporal patterns.
TMMoE \cite{yuan2023tmmoe} extends this idea to multi-task vehicle trajectory and intention prediction. 
Each task employs an independent LSTM-based gating network that captures task-specific temporal dependencies and outputs dynamic expert weights using a shared softmax layer.

\subsubsection{Non-parametric Gate}
Beyond the widely adopted parametric gating networks, several novel gating mechanisms have been proposed, offering unique advantages in specific scenarios. These approaches often replace traditional learnable networks with rule-based, algorithmic, or theoretically grounded weighting schemes, enabling dynamic and interpretable expert aggregation. 
MSGNet \cite{cai2024msgnet} adopts a purely signal-driven gating mechanism, where expert weights are computed from the amplitudes of the FFT-transformed input. The gate aggregates the frequency-domain magnitudes:
\begin{align}
    g(\mathbf{x}_t) = 
    \operatorname{Avg}\!\left(
        \operatorname{Amp}\!\left(
            \operatorname{FFT}(\mathbf{x}_t)
        \right)
    \right).
\end{align}
MoE-F \cite{saqur2025moef} replaces input-driven gating with a probabilistic filtering process. Each expert maintains a Bayesian filter updated by observed losses, and the gating distribution is modeled as the posterior of a finite-state Hidden Markov process:
\begin{align}
    g_t = \pi_t = 
    \bigl(
        \Pr(\omega_t = m \mid \mathcal{H}_t)
    \bigr)_{m=1}^{M}.
\end{align}
MOOE \cite{zhao2025mooe} further extends performance-driven gating by adopting an exponentially weighted forecaster. Expert weights are updated online according to cumulative loss, favoring experts that remain accurate over time:
\begin{align}
    g_t = \pi_t =
    \bigl(\pi_t^{(m)}\bigr)_{m=1}^{M}, \qquad
    \pi_{t+1}^{(m)} = 
    \frac{
        \pi_t^{(m)} \exp(-\eta \ell_t^{(m)})
    }{
        \sum_{j=1}^{M} 
        \pi_t^{(j)} \exp(-\eta \ell_t^{(j)})
    }.
\end{align}

\subsection{Sparse Gate}
\label{sec:sparsegate}
Unlike dense gates, which assign non-zero weights to all experts, sparse gates activate only a small subset of experts based on their gating scores, as illustrated in Figure~\ref{fig:Routing}~(c).
This sparse activation substantially reduces computational cost while encouraging expert specialization, as each expert focuses on a distinct subset of inputs. Despite sharing this objective, sparse gating mechanisms differ widely in how routing scores are produced.
We therefore categorize existing sparse gates into two major families: 
(1) \textit{parametric sparse gates}, which generate expert logits through learnable networks, 
and (2) \textit{similarity-driven sparse gates}, determine expert assignments using affinity measures derived from clustering structure, prototypes, or distance-based criteria.

\subsubsection{Parametric Gate}
Among various sparse gating designs, the MLP-based gate remains the most classical and widely adopted form. 
It typically employs a lightweight feedforward network followed by a softmax and Top-$k$ operator to convert input representations into expert logits, providing a simple yet effective mechanism for adaptive expert selection, as defined in Eq.~\ref{eq:sparse-gate}.

However, sparse activation naturally gives rise to the problem of \emph{load imbalance}, where only a few experts are frequently activated while others receive little to no routing. 
This imbalance undermines both expert specialization and overall model capacity, making load regulation a central design consideration for parametric sparse gates.

To address this issue, existing models incorporate different balancing strategies, which give rise to four major categories:
(1) \textit{Without Balancing}, where no explicit balancing strategy is used;
(2) \textit{Balance with noise injection}, where stochastic perturbation of gating logits promotes diverse expert activation;
(3) \textit{Balance with load-balancing loss}, where auxiliary objectives regulate routing distributions or expert usage frequencies; and
(4) \textit{Balance with loss and noise}, which combines stochastic exploration with explicit regularization for improved stability and diversity.

\textit{(1) Without balancing.} 
Some sparse MoE models omit explicit balancing strategies and rely solely on raw gating logits for expert assignment.
While this simplifies training, it often results in expert underutilization and weaker specialization.

MoE-AMC \cite{gao2023moeamc} uses a three-layer MLP gate to distinguish high- and low-SNR conditions for expert fusion, without applying any balancing mechanism.
ARTEMIS \cite{feng2025artemis} focuses on multi-strategy decision modeling and similarly omits balancing techniques, avoiding additional regularization to preserve expert specialization under feature imbalance.
xTime \cite{li2025xtime} employs MoE to route rarity-specific experts for extreme-event prediction, but relies purely on gating scores without explicit balancing.
MGTS-Net \cite{cai2024msgnet} incorporates a MoE-based temporal module within a multimodal graph network to capture frequency–time interactions, likewise without adopting balancing techniques.

\textit{(2) Balance with noise injection.}
To encourage expert exploration, stochastic noise can be added to the gating logits before selecting the top-$k$ experts. 
The perturbed logits $g(\mathbf{x}_t; \phi) + \mathcal{R}_{\text{noise}}$ are then passed through the TopK operator:
\begin{align}
\mathcal{G}(\mathbf{x}_t; \phi)_m =
\operatorname{softmax}\!\left(
\operatorname{TopK}\!\left(g(\mathbf{x}_t; \phi) + \mathcal{R}_{\text{noise}}, k\right)
\right)_m.
\end{align}
Noise injection introduces randomness into expert selection, promoting exploration and alleviating early expert collapse.

Several models leverage this stochastic routing to improve expert utilization and achieve implicit balance.
Pathformer \cite{chen2024pathformer} introduces multi-scale expert branches selected by a noise-injected router.
Super-Linear \cite{nochumsohn2025superlinear} first pretrains frequency-specialized experts, then learns a stochastic router to dynamically allocate inputs.
TransMoE \cite{shi2025transmoe} inserts sparse MoE layers after Transformer attention, using noise-based gating for expert activation.
MoU \cite{peng2025mou} applies the same stochastic routing to coordinate mixture-of-feature-extractor and mixture-of-architecture experts.

\textit{(3) Balance with load-balancing loss.} 
Another common approach to achieve load balancing is to introduce an auxiliary loss term that penalizes uneven expert assignments, encouraging each expert to receive a comparable fraction of inputs. This objective directly regularizes the routing distribution during training, preventing expert collapse and ensuring stable convergence without altering the gating architecture. 

The original sparsely-gated MoE \cite{shazeer2017outrageously} proposes two complementary regularizers, importance loss and load loss, to encourage uniform allocation of both gating mass and expected routing counts. 
Given a batch $\mathcal{B} = \{\mathbf{x}_1, \dots, \mathbf{x}_T\}$ with $T$ tokens and $M$ experts indexed by $m = 1, \dots, M$, and gating weights $\mathcal{G}(\mathbf{x}; \phi)_m$, the total importance and expected load of expert $m$ are defined as:
\begin{align}
\text{Imp}_m = \sum_{\mathbf{x}_t \in \mathcal{B}} \mathcal{G}(\mathbf{x}_t; \phi)_m, \quad 
\text{Load}_m = \sum_{\mathbf{x}_t \in \mathcal{B}} P(\mathbf{x}_t \rightarrow m),
\end{align}
where $P(\mathbf{x}_t \rightarrow m)$ is the probability that token $\mathbf{x}_t$ is routed to expert $m$.  
The two losses minimize the coefficients of variation (CV) of these quantities:
\begin{align}
\mathcal{L}_{\text{imp}} = \Biggl(\frac{\mathrm{Std}_m[\text{Imp}_m]}{\mathrm{Mean}_m[\text{Imp}_m]}\Biggr)^2, \quad
\mathcal{L}_{\text{load}} = \Biggl(\frac{\mathrm{Std}_m[\text{Load}_m]}{\mathrm{Mean}_m[\text{Load}_m]}\Biggr)^2.
\end{align}
These formulations establish the foundation for modern load-balancing objectives and are still used in recent time-series MoE systems \cite{liu2025softshape, chen2024bpmoe, sievers2024softdense}.

Several recent time-series MoE frameworks \cite{xiaoming2025timemoe, hong2025contextst, liu2025mofetime} employ the GShard loss \cite{lepikhin2021gshard}, which simplifies the above objectives into a computationally efficient surrogate.
Denote the routing fraction $\mathcal{D}_m$ and average gating probability $\mathcal{P}_m$ as:
\begin{align}
    \mathcal{D}_m &= \frac{1}{T} \sum_{\mathbf{x}_t \in \mathcal{B}} 
    \mathbbm{1}\{\operatorname{argmax} \mathcal{G}(\mathbf{x}_t; \phi) = m\}, \\
    \mathcal{P}_m &= \frac{1}{T} \sum_{\mathbf{x}_t \in \mathcal{B}} \mathcal{G}(\mathbf{x}_t; \phi)_m,
\end{align}
and define the loss as:
\begin{align}
    \mathcal{L}_{\text{load}} = M \sum_{m=1}^{M} \mathcal{D}_m \mathcal{P}_m.
\end{align}
Minimizing $\mathcal{L}_{\text{load}}$ encourages uniform expert activation. 
At the optimum, each expert receives approximately equal routing frequency and gating weight, i.e., $\mathcal{D}_m = \mathcal{P}_m = 1/M$, yielding $\mathcal{L}_{\text{load}} = 1$.
This balance helps fully utilize model capacity and maintain stable training dynamics.

Lastly, CAPTime \cite{yao2025captime} adopts the auxiliary balancing loss introduced in Switch Transformer \cite{fedus2022switch}, which further streamlines the objective by operating solely on gating probabilities.
Let $\mathcal{P}_m$ denote the mean gating probability as above.
The auxiliary loss is defined as:
\begin{align}
    \mathcal{L}_{\text{load}} = M \sum_{m=1}^{M} \mathcal{P}_m^2.
\end{align}
This simplified formulation directly penalizes highly skewed expert weights, achieving stable balancing with the lowest computational cost among the three methods.

\textit{(4) Balance with noise and loss.}
Some MoE systems combine stochastic exploration from noise-injected gating with loss-based regularization, jointly preventing early expert collapse and achieving more stable expert utilization.

MMK \cite{han2024MMK} incorporates the Kolmogorov–Arnold Network (KAN) into a sparse MoE for multivariate forecasting, using noise-perturbed gating together with a load-balancing loss to ensure uniform expert usage.
TALON \cite{sun2025talon} adapts LLMs to time series via a heterogeneous encoder whose pattern-aware router uses noisy gating and applies both load loss and importance loss to prevent skewed token-to-expert assignments.
MixMamba \cite{alkilane2024mixmamba} integrates Mamba experts within a sparse MoE and employs noise-injected gating alongside load loss and GShard balancing loss to maintain stable routing across experts.
TimeFilter \cite{hu2025timefilter} routes spatio-temporal patches to specialized filters for forecasting, balancing expert usage via load loss and leveraging dynamic loss \cite{huang2024harder} to encourage minimal yet effective expert activation.

\subsubsection{Similarity-driven Gate}
Similarity-driven sparse gating assigns inputs to experts based on affinity measures rather than learned logits, offering a structured and interpretable alternative to parametric routers. 
Early approaches such as TFPS \cite{sun2024tfps} and MOIRAI-MoE \cite{liu2025moiraimoe} leverage clustering structure, where inputs are routed to the nearest subspace or K-means centroid using subspace affinity or Euclidean distance. These clustering-induced inductive biases naturally group temporally coherent patterns, while expert utilization is regulated either implicitly through clustering regularization or explicitly via GShard-style load balancing.
Building on this idea, prototype-based methods introduce learnable cluster representations to enhance flexibility. 
For example, TimeTracker \cite{shi2025timetracker} and ProtoN-FM \cite{gong2026bridging} compute routing weights based on the similarity between input features and prototype matrices, while routing bias or prototype orthogonality is introduced to maintain balanced expert activation and encourage prototype diversity.
Beyond clustering and prototype learning, distance-driven gating extends similarity-based routing to more general forms: FuseMoE \cite{han2024fusemoe} computes expert affinity using Laplace-style or L2 distances to improve robustness, while FT-MoE \cite{xiao2025ftmoe} uses cosine similarity combined with a learnable threshold to dynamically select the number of active experts. TimeExpert \cite{ma2025timeexpert} further refines this approach by introducing a Temporal MoE gate that ranks key-value pairs by temporally modulated query–key similarity and selects the top-$k$ relevant ones.
Together, these approaches demonstrate how similarity-driven gates introduce powerful structural inductive biases that align naturally with heterogeneous temporal patterns, enabling stable, interpretable, and task-adaptive expert specialization without relying on heavy parametric gating networks.

\subsection{Summary}
In summary, gating strategies lie at the core of MoE architectures by determining how experts are activated, weighted, and aggregated. 
Ungated models offer architectural simplicity and stable optimization but sacrifice adaptability and expert specialization due to fixed fusion schemes. 
Dense-gated models, in contrast, activate all experts simultaneously and learn continuous weighting coefficients through various parametric or similarity-driven gating networks, achieving flexible expert integration at the cost of higher computational redundancy. 
Sparse-gated models further advance this design by activating only a subset of experts, thereby improving computational efficiency and promoting specialization, while employing load-balancing techniques to ensure stable and diversified expert utilization.

\begin{figure}[t]
\centering
\includegraphics[width=1.0\linewidth]{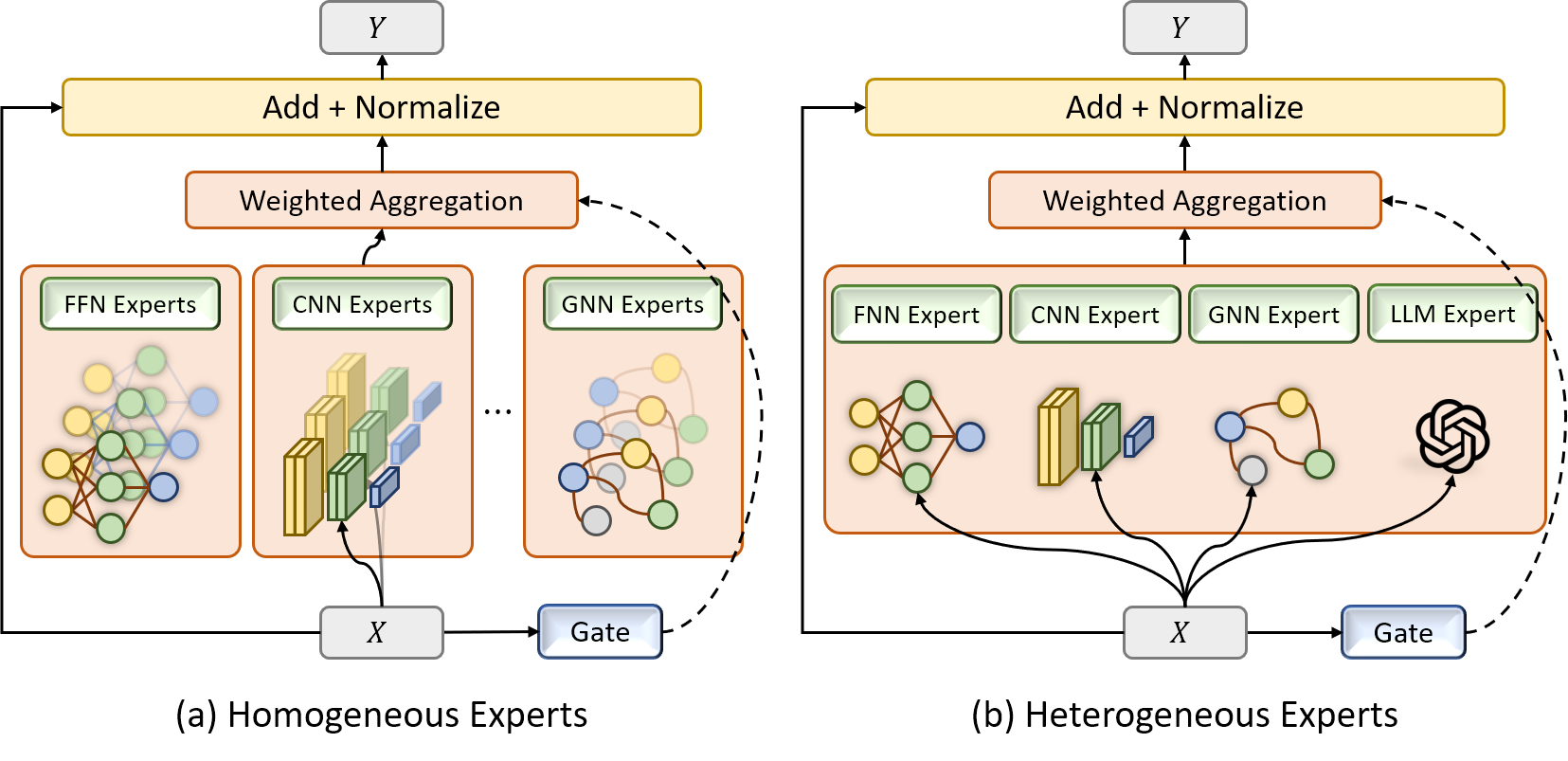}
\caption{Illustration of two types of expert configurations in the Mixture-of-Experts framework: (a) homogeneous experts, where all experts share the same architecture, and (b) heterogeneous experts, where experts employ different architectures (e.g., FNN, CNN, GNN, and LLM).}
\label{fig:Expert}
\end{figure}

\section{Expert Design}
\label{sec:expert design}
The design of experts plays a central role in determining the modeling capacity and specialization behavior of time series Mixture-of-Experts (MoE) architectures.
Recent studies in time series analysis have explored diverse expert structures tailored to different temporal dynamics, feature domains, and task objectives.
To provide a systematic view, existing approaches can be broadly categorized into (1) \textbf{Homogeneous experts}, (2) \textbf{Heterogeneous experts}, and (3) \textbf{Shared experts}. This taxonomy reflects a growing trend from uniform architectures toward multi-scale, cross-domain, and parameter-sharing designs that enhance adaptability and efficiency across varied time series scenarios.

\subsection{Homogeneous Experts}
\label{sec:homoexperts}
As illustrated in Figure~\ref{fig:Expert}~(a), homogeneous experts share an identical architectural form (e.g., FFN-only, CNN-only, or GNN-only), differing only in their parameter sets.
Each expert implements the same computation function $f(\cdot)$ but with independent parameters $\theta_m$, enabling consistent processing across experts while allowing specialization through parameter learning.
This formulation provides a clean and modular design in which expert variability arises solely from learned weights rather than architectural differences.

\subsubsection{FFN-based Experts}
Feed-Forward Network (FFN)-based experts constitute the most prevalent and foundational expert design in time-series MoE systems.
Their simplicity, efficiency, and strong representational capacity make them a natural choice for scaling model width through expert parallelism while maintaining architectural modularity.
Formally, each FFN expert applies a linear or multi-layer feedforward transformation with its own parameter set:
\begin{align}
    f_m(\mathbf{x}_t;\theta_m) = W_m\mathbf{x}_t + b_m.
\end{align}

FFN-based experts have evolved into two major categories depending on the scale and modeling ambition of the MoE system:
(1) \textit{Small-model} FFN experts designed as moderate-capacity feed-forward modules to augment backbone expressiveness and capture heterogeneous temporal patterns; and
(2) \textit{Large-model} FFN experts, which extend MoE design to large models and multimodal pretraining, enabling semantic generalization and cross-domain adaptation.

\textit{(1) Small models.}
The early works integrate FFN experts as lightweight modules into the structure of small models to enhance the expressiveness of the backbone without altering the overall network structure.
These experts typically specialize in different structural aspects of time-series data depending on how the FFN is instantiated.
For example, MoPE \cite{niu2024mope} replaces the single projection layer with multiple linear projection experts to capture inter- and intra-variate temporal dynamics, while VE \cite{wang2024ve} introduces variate-specific embedding experts to better encode multivariate correlations.
FuseMoE \cite{han2024fusemoe} incorporates modality-specific FFN experts for multimodal fusion under incomplete inputs, and FreqMoE \cite{liu2025freqmoe} decomposes signals into frequency-band experts. 
AMD \cite{hu2025amd} uses two-layer predictors for learning diverse temporal dynamics, whereas MoME \cite{li2025mome} deploys FFN experts across time, frequency, and spatial domains for comprehensive multivariate forecasting.

To model the diverse temporal structures present in real-world time series, a series of MoE frameworks introduce FFN experts that specialize in distinct regimes, scales, or contextual patterns rather than sharing a uniform modeling capacity.
TFPS \cite{sun2024tfps} constructs pattern-specific experts over temporally or spectrally clustered patches to mitigate patch-level distribution shift,
MOOE \cite{zhao2025mooe} combines offline and online experts to enable robust non-stationarity adaptation, while WaveTS-M \cite{zhou2026wavelet} and MoU \cite{peng2025mou} organize FFN experts over wavelet or semantic patch spaces to capture multi-scale variations.  
FT-MoE \cite{lee2024tfmoe} and TimeExpert \cite{ma2025timeexpert} further introduce task- and context-aware experts, enabling specialized processing under different fault conditions.  
Finally, MGTS-Net \cite{hao2025mgts} and LeMoLE \cite{zhang2024lemole} employ short-, mid-, and long-horizon forecasting experts to specialize across distinct predictive ranges, enhancing temporal resolution and robustness.

Finally, FFN-based MoE designs have also been adapted to domain-specific downstream applications.  
Examples include federated load/photovoltaic forecasting with client-adaptive experts \cite{sievers2024softdense}, oil-well yield prediction with sparse FFN experts embedded in a Transformer backbone \cite{shi2025transmoe}, and financial forecasting systems leveraging MLP experts for volatility estimation and trend prediction \cite{yumlu2003financial, ebrahimpour2011anfis}.  
Together, these architectures demonstrate the versatility of small-model FFN experts in capturing diverse temporal patterns while maintaining computational efficiency.

\textit{(2) Large models.}
Driven by the strong contextual reasoning capabilities of LLMs, recent methods increasingly integrate LLMs into time-series forecasting systems. However, LLMs inherently lack temporal inductive bias, and fine-tuning them is computationally expensive. Thus, MoE-based FFN experts are introduced as lightweight domain-adaptation modules that inject temporal priors and specialize LLM representations for time-series dynamics \cite{fan2025smetimes, yao2025captime}.
In parallel, several time-series foundation models incorporate FFN-based MoE architectures to cope with the extreme heterogeneity and multi-regime nature of temporal data across diverse domains \cite{liu2025moiraimoe, xiaoming2025timemoe}. By enabling token-level specialization while keeping computation nearly constant, these models substantially increase representational capacity and provide a scalable pathway toward large-scale time-series foundation modeling.
Finally, ULoRA-MoE \cite{wuuncertaintyuloramoe} complements these efforts from the fine-tuning perspective, incorporating uncertainty-aware LoRA-based experts to enhance robustness and calibration for anomaly detection under data contamination.

\subsubsection{LSTM-based Experts}
LSTM-based experts \cite{hochreiter1997lstm} leverage recurrent gating to capture temporal dependencies and remain effective for modeling sequential patterns within MoE systems. A typical formulation updates each expert’s hidden state as:
\begin{align}
    h_{t,m} &= \operatorname{LSTM}_m(\mathbf{x}_t, h_{t-1,m}), \\
    f_m(\mathbf{x}_t;\theta_m) &= W_m h_{t,m} + b_m.
\end{align}

In the context of energy forecasting, \cite{sievers2024softdense} embeds LSTM modules as independent experts within a MoE framework, enabling the model to learn distinct temporal dynamics for load, photovoltaic, and prosumption prediction.
For vehicle trajectory and driving intention prediction, TMMOE \cite{yuan2023tmmoe} employs task-specialized LSTM experts that capture heterogeneous motion behaviors across multiple tasks.
In the financial domain, DeepUnifiedMom \cite{ong2024deepunifiedmom} constructs LSTM experts across fast-, medium-, and slow-momentum prediction tasks, allowing the model to capture temporal dependencies at various time horizons in portfolio construction.

\subsubsection{Attention-based Experts}
Attention-based architectures \cite{vaswani2017attention} have become essential in time series modeling, effectively capturing long-range dependencies and multi-scale temporal patterns. When instantiated as experts within MoE frameworks, they typically operate by computing query–key affinities and using the resulting attention weights to aggregate value representations:
\begin{align}
    f_m(\mathbf{x}_t;\theta_m) = \operatorname{softmax}(\frac{q_m^T k_m}{\sqrt{d}})v_m.
\end{align}
where $q_m, k_m, v_m$ denote the expert-specific query, key, and value projections of the input.

Pathformer \cite{chen2024pathformer} adopts multi-scale Transformer experts, where each expert applies dual attention over temporal patches at different resolutions. This enables specialization in capturing either global correlations or fine-grained local variations, supporting robust general-purpose forecasting.
TITAN \cite{wang2024titan} employs attention-based experts, where functional specialization into sequence-centric and variable-centric subnetworks enables fine-grained temporal and spatial dependency modeling for traffic flow prediction
DynamicFusion \cite{dong2025dfmh} employs modality-specific attention experts to encode price movements, financial news, and investor sentiments through interaction attention mechanisms for financial forecasting.
POND \cite{wang2024pond} introduces domain-specific conditional prompt experts, each utilizing attention-based encoding to capture domain-dependent temporal features for multi-source time series domain adaptation. 
Graph-MoE \cite{huang2025graphmoe} incorporates graph-attention experts that fuse hierarchical multi-layer graph representations into entity embeddings, enabling adaptive modeling of structural and temporal dependencies essential for anomaly detection.

\subsubsection{Mamba-based Experts}
The emergence of Mamba \cite{gu2024mamba} has introduced state-space models (SSMs) with linear-time complexity, strong long-context modeling, and stable gradient behavior, making them attractive expert candidates for MoE-based time-series systems. A typical SSM-based expert can be written as:
\begin{align}
    f_m(\mathbf{x}_t;\theta_m) &= \operatorname{SSM}_m(\mathbf{x}_t)=\mathbf{C}_m^T\mathbf{s}_{t,m},\\
    \mathbf{s}_{t,m} &= \mathbf{A}_m\mathbf{s}_{t-1,m}+\mathbf{B}_m\mathbf{x}_t.
\end{align}

Specifically, MixMamba \cite{alkilane2024mixmamba} employs a pool of specialized Mamba-based experts that leverage the linear-time state-space formulation to model diverse and non-stationary temporal dynamics across heterogeneous time-series domains.
Similarly, MambaMoE \cite{wang2025mambamoe} integrates an MoE architecture built upon SSMs, where each expert specializes in distinct market behaviors and long-range temporal dependencies, thereby improving forecasting performance in financial time series.

\subsubsection{Other Experts}
Beyond the above expert types, several time-series MoE frameworks adopt alternative expert formulations that leverage diverse inductive biases from graph modeling, convolution, pooling, generative reconstruction, and continuous function approximation. These designs showcase the breadth and flexibility of expert specialization in MoE systems.

Traffic prediction systems \cite{chattopadhyay2022mixture} utilize multiple STGCN experts to enhance spatio-temporal modeling stability.
More recently, MSGNet \cite{cai2024msgnet} and TimeFilter \cite{hu2025timefilter} employ GNN-based experts that perform patch-specific spatial–temporal filtration, selectively preserving the most informative local correlations for complex multivariate sequences.

In addition, CoST \cite{woo2022cost} constructs each expert as a 1D autoregressive convolution with different kernel sizes, enabling multi-scale trend decomposition.
Pooling-based architectures such as TCNet \cite{zhang2023TCNet} treat average pooling at different scales as experts to capture coarse-to-fine temporal structures efficiently. TFMoE \cite{lee2024tfmoe} introduces VAE-based experts, where each expert consists of a reconstructor–predictor pair that models sensor-group dynamics and adapts to evolving traffic conditions.
KAN \cite{liu2025kan} provides a continuous and expressive function representation, and MMK \cite{han2024MMK} extends this design by constructing a mixture of KAN-based experts, where each expert focuses on modeling different subsets of variables through non-linear functional transformations.

\subsubsection{Custom Experts}
Beyond the above canonical expert families, several recent studies design custom experts whose architectures encode task-specific, cross-domain, or representation-level inductive biases.
These methods can be broadly grouped into three categories:
\textit{(1) Domain-specific experts}, which encode priors tailored to particular applications;
\textit{(2) Cross-domain modeling experts}, which enable generalization across different dynamical regimes or pretraining–finetuning gaps; and
\textit{(3) Representation-oriented experts}, which modify data transformations to improve downstream modeling robustness.

\textit{(1) Domain-specific experts.}
CAD \cite{si2023cad} introduces conflict-aware, metric-specific experts, each consisting of one convolutional layer and two feed-forward layers to enhance feature extraction capabilities in multivariate anomaly detection.
The interpretable MoE \cite{ke2024interpretable} focuses on traffic speed prediction, constructing two Temporal Fusion Transformer experts to separately capture recurrent (periodic) and non-recurrent (incident-driven) temporal behaviors.
Similarly, xTime \cite{li2025xtime} employs a homogeneous expert fusion framework for extreme event forecasting, where experts trained at different event rarity levels are dynamically fused to enhance rare-event prediction in climate and healthcare.

\textit{(2) Cross-domain modeling experts.}
DynaMix \cite{hemmer2026true} employs ALRNN-based experts \cite{brenner2024almostlinear}, each capturing a distinct dynamical regime, to achieve zero-shot generalization and long-term stability across diverse dynamical system.
Similarly, MoFE-Time \cite{liu2025mofetime} integrates Frequency-Time Cell experts, leveraging MoE routing to construct multi-dimensional sparse representations and enable cross-domain knowledge transfer from pretraining to finetuning.

\textit{(3) Representation-oriented experts.}
PTN \cite{yang2025PTN} couples a learnable data transformation expert with a co-trained predictor to jointly enhance fidelity and predictive accuracy, while ProtoNorm \cite{gong2026bridging} replaces conventional LayerNorm with prototype-guided normalization experts, where each expert corresponds to a learned prototype cluster, effectively mitigating domain shifts during time-series foundation model pretraining. 
Frequency- and function-based experts are further investigated in FEDformer \cite{zhou2022fedformer} and FiLM \cite{zhou2022film}, which replace traditional attention with spectral decomposition or functional memory projections to efficiently model long-term dependencies.
Similarly, MOE-KAN \cite{peng2024moekan} adopts a functional MoE formulation, where each expert corresponds to an average filter at a different temporal scale, enabling interpretable trend–residual decomposition in server performance forecasting.

\subsection{Heterogeneous Experts}
\label{sec:heteroexperts}
In contrast, as shown in Figure~\ref{fig:Expert}~(b), heterogeneous experts employ diverse architectural designs, enabling each expert to specialize in distinct temporal or frequency characteristics, feature correlations, or task-specific objectives.
For example, one expert may capture local temporal dependencies using convolutional structures, while another models long-range relationships through Transformer blocks or recurrent modules.
As highlighted by \cite{bourgin2021maes, ernie2025technicalreport}, such architectural heterogeneity allows the MoE to better accommodate diverse and evolving system dynamics, further enhancing its adaptability to non-stationary environments. 

\subsubsection{Statistical – Neural Model Mixtures}
Early heterogeneous MoE designs combine statistical and neural modeling paradigms to exploit their complementary inductive biases: statistical models provide interpretable trend–seasonal structures, while neural networks capture nonlinear temporal dynamics.  

Let the first $M$ experts be statistical models and the remaining $N$ experts be neural networks. A unified heterogeneous MoE layer can be written as:
\begin{align}
y_t
= \sum_{m=1}^{M} \mathcal{G}_m\, 
    f^{\mathrm{stat}}_m(\mathbf{x}_t)
  + \sum_{m=M+1}^{M+N}\mathcal{G}_m\, 
    f^{\mathrm{nn}}_{m-M}(\mathbf{x}_t).
\end{align}

A statistical expert may be instantiated as:
\begin{align}
    f^{\mathrm{stat}}_m(\mathbf{x}_t) 
    = \mathcal{S}_m(z_t),
\end{align}
where $z_t$ contains lagged values or engineered statistical features and $\mathcal{S}_m$ denotes a statistical forecasting model (e.g., ARIMA, SARIMA). A neural expert follows:
\begin{align}
    f^{\mathrm{nn}}_n(\mathbf{x}_t) 
    = \mathrm{NN}_n(\mathbf{x}_t;\theta^{\mathrm{nn}}_n).
\end{align}

DYCHEM \cite{han2022dychem} exemplifies this hybrid design by combining ARIMA, dynamic linear models, Prophet, and DeepAR as heterogeneous experts within a hierarchical MoE, producing coherent point and quantile forecasts in large-scale financial applications.  
Similarly, TSEPMoE \cite{chen2023tsepmoe} blends SARIMA, SARIMAX, and P-SARIMA with neural components for electric load forecasting, demonstrating that integrating complementary statistical and neural structures improves robustness across varying temporal regimes.

\subsubsection{Neural Model Mixtures}
As deep learning matured, heterogeneous MoE structures evolved toward integrating multiple neural architectures with complementary temporal, spatial, or functional biases.
Let $\mathcal{A}_s$ denote the neural architecture family of type $s$ (e.g., LSTM, Transformer, CNN). The MoE output is:
\begin{align}
y_t
= \sum_{s=1}^{S} \sum_{m=1}^{M}
  \mathcal{G}^{(s)}_m\,
  \mathcal{A}_s(\mathbf{x}_t; \theta^{(s)}_m).
\end{align}

\textit{(1) General-purpose heterogeneous neural experts.}
TALON \cite{sun2025talon} employs a heterogeneous temporal encoder that partitions time series into coherent segments and assigns Linear, CNN, and LSTM experts to capture diverse local dynamics.
MoME \cite{li2025mome} integrates linear trend-extraction modules with Transformer-based spatial–frequency experts to jointly model temporal, spatial, and spectral dependencies in multivariate long-term forecasting.
Following a similar temporal–frequency fusion paradigm, MGTS-Net \cite{hao2025mgts} constructs each expert using a Frequency–Time Cell (FTC) paired with a FFN, allowing dynamic weighting between time- and frequency-domain representations to capture joint temporal–spectral structure.

\textit{(2) Application-specific heterogeneous neural experts.}
IMMoE \cite{tong2023immoe} combines CNN, gated recurrent unit (GRU), and autoregressive experts within a multi-gate MoE framework to capture both short- and long-term dynamics in gas load forecasting.
Garg et al. \cite{garg2024time} design a hybrid MoE for air cargo demand forecasting, mixing Deep Neural Network, N-BEATS, and Temporal Fusion Transformer experts to balance interpretability and accuracy under volatile market conditions.
Vallarino et al. \cite{vallarino2024dynamic} fuse LSTM and linear experts for stock prediction, where nonlinear volatility patterns and stable market trends are adaptively balanced across market regimes.
MoE-AMC \cite{gao2023moeamc} tailors Transformer-based (LSRM) and ResNet-based (HSRM) experts to low- and high-SNR conditions in automatic modulation classification, achieving robust performance across noise environments.
BP-MoE \cite{chen2024bpmoe} introduces behavior-pattern–aware experts composed of GRU, Graph Attention Network, and MLP encoders to jointly model long-term temporal memory, spatial dependencies, and short-term behavioral dynamics for temporal graph tasks such as link prediction and node classification.

\subsubsection{Foundation Model Mixtures}
A growing line of heterogeneous MoE systems incorporates LLMs as experts, leveraging their strong semantic reasoning capabilities to complement neural or statistical forecasters. 
In this setting, an expert $f_m(\cdot;\theta_m)$ in Eq.~\ref{eq:dense_output} and~\ref{eq:sparse_output} may correspond to a full-fledged pretrained LLM, possibly with different sizes or with distinct fine-tuned or LoRA-adapted parameters, while other experts capture numerical or dynamical temporal patterns.
These architectures thus form mixtures where at least one expert is an LLM, enabling cross-modal or cross-domain generalization.

MoE-F \cite{saqur2025moef} directly treats multiple pretrained LLMs as independent forecasting experts, each providing distinct semantic and contextual priors for financial time series. 
VPF-MoE \cite{zhao2025vpfmoe} constructs a hybrid MoE in which LLM experts supply high-level semantic reasoning while neural time-series experts model fine-grained temporal dynamics, jointly improving robustness in vegetable price prediction. 
MSE-ITT \cite{koval2025MSEITT} adopts a modality-specific MoE where a pretrained language model acts as the text expert and a Transformer-based module acts as the time-series expert, enabling coherent multimodal forecasting from interleaved news and financial sequences.

\begin{figure}[t]
    \centering
    \includegraphics[width=0.9\linewidth]{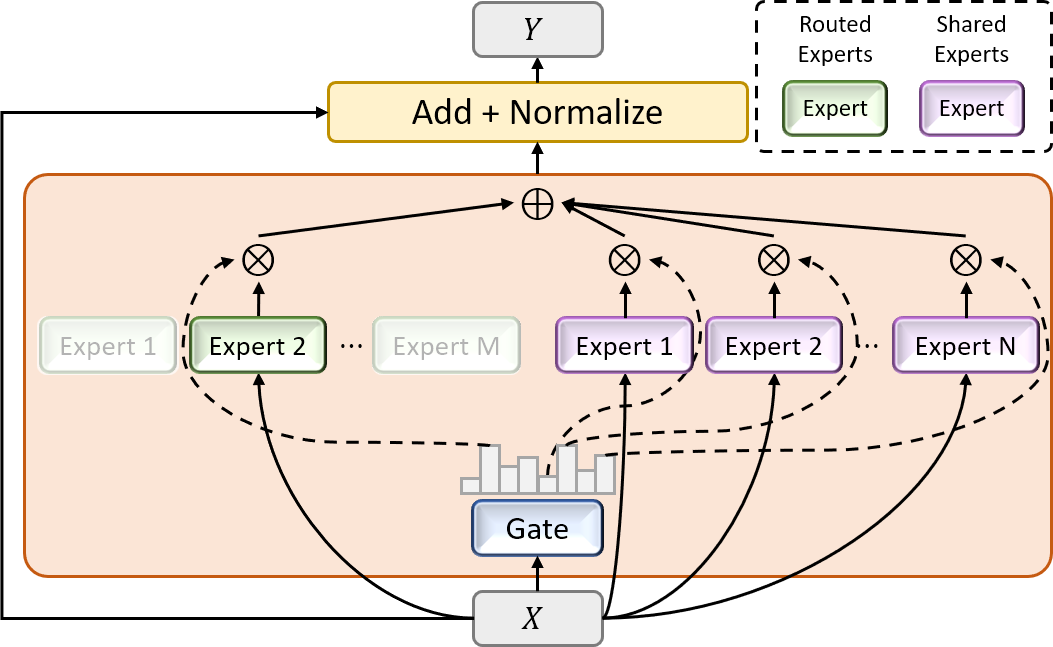}
    \caption{Illustration of shared experts that reuse common parameters or submodules to enhance efficiency and cross-task generalization.}
    \label{fig:shared}
\end{figure}

\subsection{Shared Experts}
\label{sec:sharedexperts}
Shared-expert architectures aim to balance expressiveness and efficiency by partially sharing parameters across experts.
Instead of maintaining fully independent networks, these designs introduce shared submodules that encode general temporal priors while retaining routed experts for task- or domain-specific specialization.
This hybrid structure mitigates expert underutilization, reduces computational overhead, and encourages knowledge transfer across related time-series domains.

As shown in Figure~\ref{fig:shared}, a single gating network scores all experts, with top-$k$ sparse activation applied to the first $M$ routed experts and dense normalization applied to the remaining $S$ shared experts. The final output is:
\begin{align}
    y =
    \sum_{m=1}^{M} \tilde{\mathcal{G}}_m f_m(\mathbf{x})
    +
    \sum_{m=M+1}^{M+S} \bar{\mathcal{G}}_m f_m(\mathbf{x}),
\end{align}
where $\tilde{\mathcal{G}}$ applies top-$k$ masking to the routed experts and $\bar{\mathcal{G}}$ applies dense normalization to the shared experts.
 
Early compact models explore lightweight shared structures.
MoLE \cite{ni2024mole} shares linear components across experts to enhance periodicity modeling and long-horizon forecasting efficiency,
while SoftShape \cite{liu2025softshape} combines class-specific modules with a shared expert that captures inter-shape dependencies for time-series classification.
AME \cite{zhang2024ame} extends this idea to multivariate imputation by embedding shared experts within an attention backbone to model nonlinear inter-variable correlations.

Building upon these compact architectures, large-scale time series foundation models further generalize shared-expert principles.
Time-MoE \cite{xiaoming2025timemoe} pretrains sparse shared Transformer experts across multiple domains, enabling scalable parameter growth without proportional computation, and
Time Tracker \cite{shi2025timetracker} reuses shared Transformer blocks to unify channel-independent and channel-mixed modeling in both pretraining and finetuning.

To further promote generalization, shared experts have also proven effective for cross-domain and cross-task transfer.
ContexTST \cite{hong2025contextst} incorporates shared context-aware Transformer experts that generalize temporal features while adapting domain semantics via contextual anchors.
Similarly, PatchMoE \cite{wu2025PatchMoE} employs shared hierarchical experts spanning temporal and channel dimensions, facilitating knowledge transfer across forecasting, classification, and imputation.

Finally, in specialized applications such as autonomous driving, ARTEMIS \cite{feng2025artemis} utilizes shared autoregressive experts to maintain consistent temporal reasoning and improve trajectory generation robustness.

\subsection{Summary}
In summary, the design of experts in time series MoE architectures reflects a fundamental balance among representation diversity, computational efficiency, and knowledge sharing.
Homogeneous experts ensure architectural consistency and stable optimization but are often limited in handling the multifaceted dynamics of real-world time series.
Heterogeneous experts introduce architectural variety that empowers the model to capture complex temporal dependencies and cross-domain heterogeneity, albeit at the expense of higher training and routing complexity.
Shared experts, on the other hand, strike a balance between flexibility and efficiency by enabling partial parameter reuse and domain transfer.
Collectively, these paradigms represent a gradual evolution from uniform, static expert modeling toward adaptive, modular, and task-aware expert systems that form the foundation for more advanced routing and specialization mechanisms in time series MoE frameworks.

\section{Discussion and Future Directions}
\label{sec:future}
Our earlier analysis of routing and expert architectures reveals that existing time-series MoE systems remain largely inherited from vision and NLP. Their advantages mostly reflect the intrinsic capabilities of MoE, not designs tailored to temporal structure or dynamics. This misalignment calls for a fundamental rethinking of MoE in the context of time-series modeling.

Looking forward, we argue that the next generation of time-series MoE should transition from static, architecture-centric formulations toward systems that are structure-aware, self-evolving, capable of reasoning, and ultimately endowed with agentic decision-making capabilities. Guided by this perspective, we delineate a forward-looking research roadmap centered on three tightly coupled themes: \emph{structural design}, \emph{adaptive intelligence}, and \emph{trustworthy deployment}.

Collectively, these directions chart a coherent trajectory toward a more autonomous, efficient, and responsible paradigm of MoE-driven time-series intelligence, extending the role of MoE from specialized architectural components to foundational mechanisms for future temporal AI systems.

\subsection{Structural Design}
\label{sec:structural}
\subsubsection{Modeling Channel-wise Relational Patterns}
\label{sec:channel}
The informative content of time series arises from two complementary aspects: temporal evolution and inter-channel dependency. 
However, most current MoE-based models emphasize temporal dynamics while overlooking variable interactions that encode structural correlations among channels. 
Recent work \cite{chen2024similarity} shows that channels often exhibit distinct degrees of dependency when grouped dynamically by intrinsic similarity, suggesting that variables convey heterogeneous relational patterns. 
A promising future direction is to design channel-wise MoE frameworks that explicitly model inter-variable dependencies through specialized experts, thereby enabling adaptive relation mining and more expressive multivariate representations.

\subsubsection{Automated Temporal Expert Discovery and Configuration}
Existing MoE architectures typically rely on manually defined numbers and configurations of experts, which limits their scalability and adaptability across datasets and domains \cite{sun2024tfps}. 
Future research should explore automated mechanisms that dynamically determine the number, type, and specialization level of experts based on data complexity or distributional shifts. 
Data-driven and Bayesian formulations for expert discovery could allow MoE systems to self-organize their expert pool, achieving more efficient specialization and generalization without extensive manual tuning.
Building upon automated specialization, improving efficiency and scalability becomes essential for large-scale deployment.

\subsubsection{Efficient and Scalable Architectures for Time Series}
Although MoE architectures enhance modeling capacity, they also incur significant computational and memory costs, particularly for dense temporal data and foundation-scale models. 
Enhancing efficiency thus remains an important challenge for large-scale time series MoE. 
Potential solutions include sparse expert activation and routing, parameter sharing across related experts, and expert distillation to compress large ensembles. 
Such advances will improve scalability and energy efficiency, facilitating the deployment of MoE systems in real-time, resource-constrained, and streaming time series environments.

\subsection{Adaptive Intelligence}
\label{sec:adaptive}
\subsubsection{Toward Self-evolving MoE for Non-stationary Time Series}
Real-world time series are inherently non-stationary, with evolving distributions and emerging temporal patterns over time. 
Most MoE frameworks, however, are trained in a static fashion and cannot adapt once deployed, limiting their robustness to distributional shifts. 
A key research direction is to develop continual or lifelong MoE architectures that can dynamically expand, prune, or update experts as data distributions evolve. 
Such systems could maintain a balance between stability and plasticity by retaining prior knowledge while incorporating new experts to model unseen patterns, enabling sustainable forecasting under long-term temporal evolution.
Beyond continual adaptation, the next frontier lies in enhancing the cognitive capability of experts through explicit reasoning.

\subsubsection{Temporal Reasoning-centric MoE}
While current MoE frameworks excel at pattern recognition, their reasoning capability, including how and why experts make decisions, remains underdeveloped.
Future research should move beyond mapping inputs to outputs toward equipping experts with explicit reasoning mechanisms \cite{chang2025survey}. 
This could involve endowing each expert with structured inference modules, causal abstraction, or symbolic reasoning layers that enable multi-step, explainable decision chains rather than direct correlations. 
Moreover, reasoning-aware gating strategies could dynamically select or compose experts based on logical dependencies, temporal causality, or goal-directed planning. 
By integrating reasoning into the expert architecture, MoE systems may evolve from reactive predictors to analytical agents capable of understanding, explaining, and generalizing across diverse time series scenarios.
Equipped with reasoning ability, MoE systems can further evolve toward agentic intelligence, where experts operate autonomously, interact with one another, and self-organize.

\subsubsection{Agentic MoE for Time-series Intelligence}
The next frontier of MoE research lies in transforming passive expert modules into autonomous, interacting agents \cite{tang2025unveiling}. 
In this agentic paradigm, each expert operates as an intelligent entity endowed with perception, reasoning, and decision capabilities, engaging in negotiation, collaboration, or competition within a multi-agent ecosystem \cite{huang2025many}. 
Such an Agent-MoE framework enables dynamic task allocation, adaptive routing, and continual co-evolution of experts through communication protocols or reinforcement learning signals, fostering emergent coordination and self-organization. 
Future investigations may explore multi-agent communication, co-distillation, and competition-driven specialization to achieve a balance between diversity and coherence, ultimately guiding time-series MoE systems toward more adaptive, generalizable, and self-governing intelligence.

\subsection{Trustworthy Deployment}
\label{sec:trustworthy}
\subsubsection{Privacy-preserving Multi-entity Time-series MoE}
Trustworthy deployment is particularly critical in time-series applications, as sequential data often encode long-term behavioral patterns and operational routines that attackers may exploit through cross-modal or structure-aware attacks \cite{gong2024cross, gong2022person}. 
Beyond generic model-level risks, time-series MoE faces privacy challenges arising from multi-entity deployment (e.g., across hospitals or devices) and the temporal nature of sequential data \cite{lin2024backtime, dong2025trojantime, schuchardt2025privacy}. 
In such settings, different experts may originate from distinct organizations or edge nodes trained on proprietary data streams; consequently, routing dynamics and expert parameters can inadvertently reveal institution-specific statistics, operational cycles, or user routines over time. 
A promising direction for future work is to develop privacy-aware expert orchestration in time-series MoE systems by leveraging techniques such as DP-based filtering or denoising mechanisms \cite{yan2024dpc} to mitigate cross-entity leakage while preserving utility.
Such designs would directly address the realities of cross-hospital and cross-device collaboration while maintaining utility, thereby enabling trustworthy and regulation-aligned deployment of time-series MoE in practice.

\section{Conclusion}
\label{sec:conclusion}
This article presents a comprehensive review of mixture-of-experts (MoE) methods for time series modeling and proposes a new taxonomy that systematically organizes existing works from two complementary perspectives: \emph{gating strategy} and \emph{expert design}. 
Based on these perspectives, we further categorize current approaches into six representative subcategories, covering diverse architectures along the dimensions of routing strategy and expert design.
In addition, we discuss several promising research directions to guide the future development of time series MoE frameworks. 
We believe that this review fills an important gap in the literature by offering both a structured understanding and forward-looking insights, thereby fostering deeper exploration and innovation in MoE-based time series modeling.

\bibliographystyle{IEEEtran}
\bibliography{reference}

\end{document}